\newif\ifsuppl \suppltrue  

\newif\ifsmallfonts \smallfontstrue

\documentclass[letterpaper, 10 pt, conference]{ieeeconf}  
\usepackage{microtype}
\usepackage{booktabs} 

\usepackage{silence}
\usepackage{hyperref}
\usepackage{amsmath,bm}

\usepackage{graphicx}
\usepackage{float}
\newfloat{codefloat}{tbp}{lop}

\usepackage{placeins}
\usepackage{caption}
\usepackage{subcaption}  
\usepackage{xspace}
\usepackage{booktabs} 
\usepackage{amsmath}
\usepackage[most]{tcolorbox}
\usepackage{tcolorbox}
\usepackage[T1]{fontenc}
\usepackage{amsfonts,amssymb}
\usepackage{bigdelim}
\usepackage{tabularx}
\usepackage{bbm}
\usepackage{calc}
\usepackage{footmisc}
\usepackage{mathrsfs}
\usepackage{mathtools}
\usepackage{stfloats}  
\usepackage{setspace}
\usepackage{tikz}
\usepackage{textgreek}
\usepackage[noend]{algpseudocode}
\usepackage{makecell}
\usetikzlibrary{tikzmark}
\usetikzlibrary{arrows.meta}
\usetikzlibrary{bending}
\usetikzlibrary{calc,positioning,quotes}
\usetikzlibrary{matrix,shapes.arrows}

\makeatletter
\DeclareRobustCommand\onedot{\futurelet\@let@token\@onedot}
\def\@onedot{\ifx\@let@token.\else.\null\fi\xspace}
\def\eg{\textit{e.g}\onedot}

\def\ie{\textit{i.e}\onedot}

\def\etc{\textit{etc}\onedot}

\makeatother

\newcommand{\supplsection}[3]{
    \ifsuppl%
        \section{#1}
        \label{#2}
        #3
    \else%
        \par\refstepcounter{section}%
        \sectionmark{#1}%
        \label{#2}
    \fi%
}

\newcommand{\supplheader}{
    \renewcommand{\figurename}{Supplementary Figure}
    \renewcommand{\tablename}{Supplementary Table}
    \renewcommand{\thesection}{S\arabic{section}}
    \renewcommand{\thefigure}{S\arabic{figure}}
    \renewcommand{\thetable}{S\arabic{table}}
    \setcounter{section}{0}
    \setcounter{figure}{0}
    \setcounter{table}{0}
    \ifsuppl%
        \FloatBarrier
        \clearpage
        \twocolumn[
        \centerline{\textbf{\LARGE Supplementary Material}}
        \vspace{30pt}
        ]
    \fi%
}


\def\codefont{\fontfamily{lmtt}\selectfont}
\newcommand{\textcode}[1]{{\normalfont\codefont #1}}

\newcounter{codelinecounter}[section]
\definecolor{codebackground}{rgb}{1.0,1.0,1.0}
\definecolor{codebackgroundprogress}{rgb}{1.0,0.99,0.98}
\definecolor{codeframe}{rgb}{0.8,0.8,0.8}
\definecolor{codegreen}{rgb}{0,0.4,0}
\definecolor{codeblue}{rgb}{0.25,0.25,0.75}
\definecolor{codegray}{rgb}{0.5,0.5,0.5}
\def\codefontsize{\fontsize{8.5}{9}\selectfont}  
\newcommand{\codeline}[1]{{#1\par}}
\newcommand{\codelinenum}{\stepcounter{codelinecounter} {\color{codegray} \ifnum\value{codelinecounter}<10 0\fi\arabic{codelinecounter}}\ }

\newcommand{\codecomment}[1]{{\color{codegray} \# #1}}

\newcommand{\codedef}[1]{{\color{codegreen} #1}}
\newcommand{\codereturn}[1]{{\color{codegreen} #1}}
\newcommand{\codetab}{~~}

\newenvironment{code}[3]{  
    \setcounter{codelinecounter}{0}
    \begin{tcolorbox}[
        width=#1,height=#2,
        valign=center,left=0pt,right=0pt,top=0pt,bottom=0pt,
        colback=#3,colframe=codeframe,boxrule=0.5pt,arc=0pt]
    \codefont
    \codefontsize
}{ 
    \end{tcolorbox}
}

\usepackage{todonotes}

\makeatletter
\newcommand\thefontsize{{[[[This font is \f@size pt]]]}}
\makeatother

\usepackage[english]{babel}
\usepackage{blindtext}
\usepackage{lipsum}

\title{\LARGE \bf
Discovering Adaptable Symbolic Algorithms from Scratch
}

\author{Stephen Kelly$^{1,4}$, Daniel S. Park$^{1}$, Xingyou Song$^{1,2}$, Mitchell McIntire$^{3}$, Pranav Nashikkar$^{3}$, \\ Ritam Guha$^{5}$, Wolfgang Banzhaf$^{5}$, Kalyanmoy Deb$^{5}$, Vishnu Naresh Boddeti$^{5}$, Jie Tan$^{1,2}$, Esteban Real$^{1,2}$ \\
{\small $^{1}$Google Research, $^{2}$ Google DeepMind, $^{3}$Google, $^{4}$McMaster University, $^{5}$Michigan State University}
}

\begin{document}

\maketitle
\thispagestyle{empty}
\pagestyle{empty}


\begin{abstract}
Autonomous robots deployed in the real world will need control policies that rapidly adapt to environmental changes. To this end, we propose AutoRobotics-Zero (ARZ), a method based on AutoML-Zero that discovers zero-shot adaptable policies from scratch.
In contrast to neural network adaptation policies, where only model parameters are optimized, ARZ can build control algorithms with the full expressive power of a linear register machine. We evolve modular policies that tune their model parameters \textit{and} alter their inference algorithm on-the-fly to adapt to sudden environmental changes.
We demonstrate our method on a realistic simulated quadruped robot, for which we evolve safe control policies that avoid falling when individual limbs suddenly break. This is a challenging task in which two popular neural network baselines fail.
Finally, we conduct a detailed analysis of our method on a novel and challenging non-stationary control task dubbed Cataclysmic Cartpole. Results confirm our findings that ARZ is significantly more robust to sudden environmental changes and can build simple, interpretable control policies.
\end{abstract}

\let\thefootnote\relax\footnotetext{Published and Best Overall Paper Finalist at IROS 2023.}
\let\thefootnote\relax\footnotetext{Videos: \textcolor{blue}{\url{https://youtu.be/sEFP1Hay4nE}}}
\let\thefootnote\relax\footnotetext{Correspondence: \url{spkelly@mcmaster.ca}}

\section{Introduction}
\label{intro_sec}
Robots deployed in the real world will inevitably face many environmental changes. For example, robots' internal conditions, such as battery levels and physical wear-and-tear, and external conditions, such as new terrain or obstacles, imply that the system's dynamics are non-stationary. In these situations, a static controller that always maps the same state to the same action is rarely optimal. Robots must be capable of continuously adapting their control policy in response to the changing environment. To achieve this capability, they must recognize a change in the environment without an external cue, purely by observing how actions change the system state over time, and update their control in response. Recurrent deep neural networks are a popular policy representation to support fast adaptation. However, they are often (1) monolithic, which leads to the \textit{distraction dilemma} when attempting to learn policies that are robust to multiple dissimilar environmental physics \cite{hasselt2019, kelly21}; (2) overparameterized, which can lead to poor generalization and long inference time; and (3) difficult to interpret. Ideally, we would like to find a policy that can express multiple modes of behavior while still being simple and interpretable. 

\def\arzcode{
    \begin{code}{0.48\textwidth}{4.4in}{codebackground}
        \scriptsize
        \codeline{\codecomment{wX: vector memory at address X.}}
        \codeline{\codedef{def} f(x, v, i):}
        \codeline{\codetab w0 = copy(v)}
        \codeline{\codetab w0[i] = 0}
        \codeline{\codetab w1 = abs(v)}
        \codeline{\codetab w1[0] = -0.858343 * norm(w2)}
        \codeline{\codetab w2 = w0 * w0}
        \codeline{\codetab \codereturn{return} log(x), w1}
        \codeline{~}
        \codeline{\codecomment{sX: scalar memory at address X.}}
        \codeline{\codecomment{vX: vector memory at address X.}}
        \codeline{\codecomment{obs, action: observation and action vectors.}}
        \codeline{\codedef{def} GetAction(obs, action):}
        \codeline{\codetab if s13 < s15: s5 = -0.920261 * s15}
        \codeline{\codetab {\color{red} if s15 < s12:} s8, v14, i13 = 0, min(v8, sqrt(min(0, v3))), -1}
        \codeline{\codetab if s1 < s7: s7, action = f(s12, v0, i8)}
        \codeline{\codetab action = heaviside(v12)}
        \codeline{\codetab if s13 < s2: s15, v3 = f(s10, v7, i2)}
        \codeline{\codetab if s2 < s0: s11, v9, i13 = 0, 0, -1}
        \codeline{\codetab s7 = arcsin(s15)}
        \codeline{\codetab if s1 < s13: s3 =  -0.920261 * s13}
        \codeline{\codetab {\color{red} s12 = dot(v3, obs)}}
        \codeline{\codetab s1, s3, s15  = maximum(s3, s5), cos(s3), 0.947679 * s2}
        \codeline{\codetab if s2 < s8: s5, v13, i5 = 0, min(v3, sqrt(min(0, v13))), -1}
        \codeline{\codetab if s6 < s0: s15, v9, i11 = 0, 0, -1}
        \codeline{\codetab if s2 < s3: s2, v7 = f3(s8, v12, i1)}
        \codeline{\codetab if s1 < s6: s13, v14, i3 = 0, min(v8, sqrt(min(0, v0))), -1}
        \codeline{\codetab if s13 < s2: s7 = -0.920261 * s2}
        \codeline{\codetab if s0 < s1: s3 = -0.920261 * s1}
        \codeline{\codetab if s7 < s1: s8, action = f(s5, v15, i3) }
        \codeline{\codetab if s0 < s13: s5, v7 = f(s15, v7, i15)}
        \codeline{\codetab s2 = s10 + s3}
        \codeline{\codetab {\color{red} if s7 < s12:} s11, v13 = f(s9, v15, i5)}
        \codeline{\codetab if s4 < s11: s0, v9, i13 = 0, 0, -1}
        \codeline{\codetab s10, action[i5]  = sqrt(s7), s6}
        \codeline{\codetab if s7 < s9: s15 = 0}
        \codeline{\codetab if s14 < s11: s3 = -0.920261 * s11}
        \codeline{\codetab if s8 < s5: s10, v15, i1 = 0, min(v13, sqrt(min(0, v0))), -1}
        \codeline{\codetab \codereturn{return} action}
    \end{code}
}

\begin{figure}
    \centering
    \begin{minipage}[t]{\textwidth}
    \arzcode
    \end{minipage}

    \begin{picture}(0, 0)(-17, -260)
    \includegraphics[width=0.19\textwidth]{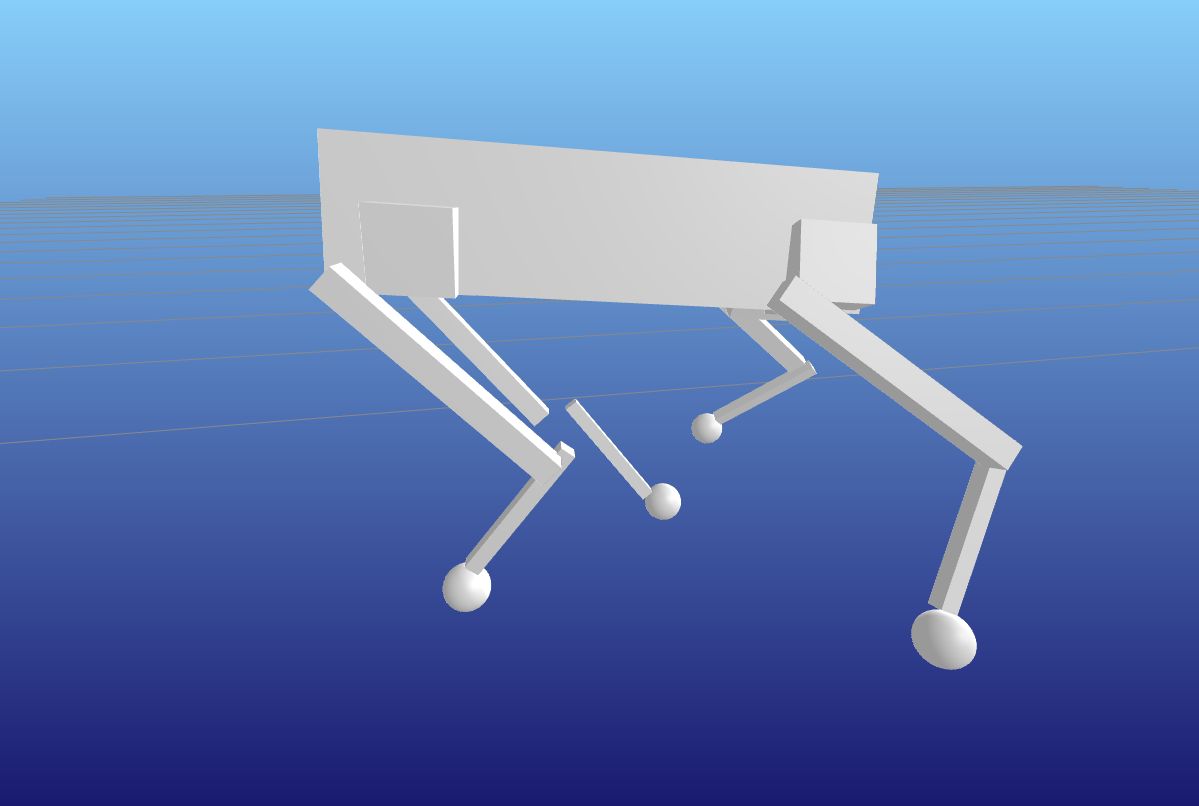}
    \end{picture}

    \caption{Automatically discovered Python code representing an adaptable policy for a realistic quadruped robot simulator (top--right inset). This evolved policy outperforms MLP and LSTM baselines when a random leg is suddenly broken at a random time. (Lines in \textcolor{red}{red} will be discussed in the text).}
    \label{quadruped_algo}
\end{figure}

We propose AutoRobotics-Zero (ARZ), a new framework based on AutoML-Zero (AMLZ) \cite{real2020automl} to specifically support the evolution of dynamic, self-modifying control policies in a realistic quadruped robot adaptation task. We represent these policies as \emph{programs} instead of neural networks and demonstrate how the adaptable policy and its initial parameters can be evolved from scratch using only basic mathematical operations as building blocks. Evolution can discover control programs that use their sensory-motor experience to fine-tune their policy parameters or alter their control logic on-the-fly while interacting with the environment. This enables the adaptive behaviors necessary to maintain near-optimal performance under changing environmental conditions.  Unlike the original AMLZ, we go beyond toy tasks by tackling the simulator for the actual Laikago robot \cite{laikagounitree}. To facilitate this, we shifted away from the supervised learning paradigm of AMLZ. We show that evolved programs can adapt during their lifetime without explicitly receiving any supervised input (such as a reward signal). Furthermore, while AMLZ relied on the hand-crafted application of three discovered functions, we allow the number of functions used in the evolved programs to be determined by the evolutionary process itself. To do this, we use conditional automatically defined functions (CADFs) and demonstrate their impact. With this approach, we find that evolved adaptable policies are significantly simpler than state-of-the-art solutions from the literature because evolutionary search begins with minimal programs and incrementally adds complexity through interaction with the task domain. Their behavior is highly interpretable as a result.

In the quadruped robot, ARZ is able to evolve adaptable policies that maintain forward locomotion and avoid falling, even when all motors on a randomly selected leg fail to generate any torque, effectively turning the leg into a passive double pendulum. In contrast, despite comprehensive hyperparameter tuning and being trained with state-of-the-art reinforcement learning methods, MLP and LSTM baselines are unable to learn robust behaviors under such challenging conditions. 

While the quadruped is a realistic complex task, simulating the real robot is time-consuming. 
Due to the lack of efficient yet challenging benchmarks for adaptive control, we created a toy adaptation task dubbed \emph{Cataclysmic Cartpole} and repeated our analysis on this task with similar findings. In both cases, we provide a detailed analysis of evolved control programs to explain how they work, something notoriously difficult with \textit{black box} neural network representations.

\renewcommand\labelitemi{\raisebox{1.5pt}{\fontsize{8}{8}$\bullet$}}

In summary, this paper develops an evolutionary method for the automated discovery of adaptable robotic policies from scratch. We applied the method to two tasks in which adaptation is critical, Quadruped Leg-Breaking and Cataclysmic Cartpole. On each task, the resulting policies:
\begin{itemize}
    \item surpass carefully-trained MLP and LSTM baselines;
    \item are represented as interpretable, symbolic programs; and
    \item use fewer parameters and operations than the baselines.
\end{itemize}
These points are demonstrated for each task in Section~\ref{results_sec}. 
\section{Related Work}
\label{related_sec}
Early demonstrations of Genetic Programming (GP) established its power to evolve optimal nonlinear control policies from scratch that were also simple and interpretable \cite{koza1990}. More recently, GP has been used to distill the behavior of complex neural network policies developed with Deep Reinforcement Learning into interpretable and explainable programs without sacrificing control quality \cite{dhebar2022}. In this work, we extend these methods to evolve programs that can change their behavior in response to a changing environment. 

We demonstrate how to automatically discover a controller that can \textit{context switch} between distinct behavior modes when it encounters diverse tasks, thus avoiding trade-offs associated with generalization across diverse environmental physics. If we can anticipate the nature of the environmental change a robot is likely to encounter, we can simulate environments \textit{similar} to the expected changes and focus on building multitask control policies \cite{kelly21, yu17}. In this case, some form of domain randomization \cite{tobin17} is typically employed to expose candidate policies to a breadth of task dynamics. However, policies trained with domain randomization often trade optimality in any particular environment dynamics for generality across a breadth of dynamics. This is the problem we aim to address with ARZ.  Unlike previous studies in learning quadruped locomotion in the presence of non-stationary morphologies (e.g., \cite{cully15}), we are specifically interested in how controllers can be automatically built from scratch without requiring any prior task decomposition or curriculum learning. This alleviates some burden on robotics engineers and reduces researcher bias toward known machine learning algorithms, opening the possibility for a complex adaptive system to discover something new. 

In addition to anticipated non-stationary dynamics, another important class of adaptation tasks in robotics is sim-to-real transfer [11], where the robot needs to adapt policies trained in simulation to unanticipated characteristics of the real-world. Successful approaches to learn adaptive policies can be categorized by three broad areas of innovation: (1) New adaptation operators that allow policies to quickly tune their model parameters within a small number of interactions \cite{xingyou20,xingyou2019,yu19,finn17}; (2) Modular policy structures that separate the policy from the adaptation algorithm and/or world model, allowing \textit{both} to be learned \cite{kumar21, najarro20, floreano2000, timothee2021}; and (3) Hierarchical methods that allow a diverse set of complete or partial behaviors to be dynamically switched in and out of use at run-time, adapting by selecting the best strategy for the current environmental situation \cite{cully15, kelly21, li2020}. These algorithmic models of behavioral plasticity, modular structure, and hierarchical representations reflect the fundamental properties of meta-learning. In nature, these properties emerged through adaptation at two timescales (evolution and lifetime learning) \cite{wang2021}. ARZ makes these two time scales explicit by implementing an evolutionary search loop that acts on a ``genome'' of code, and an evaluation that steps through an episode which is analogous to the ``lifetime'' of the robot.
\section{Methods}
\label{methods_sec}

\subsection{Algorithm Representation}\label{algo_rep_sec}
As in the original AutoML-Zero \cite{real2020automl}, policies are represented as linear register machines that act on virtual memory \cite{brameier07}. In this work, we support four types of memory: scalar, vector, matrix, and index (\eg\ \textcode{s1}, \textcode{v1}, \textcode{m1}, \textcode{i1}). Scalar, vector, and matrix memory are floating-point, while index memory stores integers.  Algorithms are composed of two core functions: \textcode{StartEpisode()} and \textcode{GetAction()}. \textcode{StartEpisode()} runs once at the start of each episode of interaction with the environment. Its sole purpose is to initialize the contents of virtual memory with evolved constants. The content of these memories at any point in time can be characterized as the control program's state. Our goal is to discover algorithms that can adapt by tuning their memory state or altering their control code on-the-fly while interacting with their environment. This adaptation, as well as the algorithm's decision-making policy, are implemented by the \textcode{GetAction()} function, in which each instruction executes a single operation (\eg \textcode{s0=s7*s1} or \textcode{s3=v1[i2]}). We define a large library of operations (Table \ref{ops_table}) and place no bounds on the complexity of programs. Evolutionary search is employed to discover what sequence of operations and associated memory addresses appear in the \textcode{GetAction()} function. 

\textit{Conditional Automatically Defined Functions:}
In addition to \textcode{StartEpisode()} and \textcode{GetAction()}, up to 6 Conditionally-invoked Automatically Defined Functions \cite{koza1994genetic} (CADFs)  may be generated in an algorithm. Each CADF represents an additional function block, itself automatically discovered, which is callable from \textcode{GetAction()}. Since each CADF is conditionally invoked, the sequence of CADFs executed at each timestep throughout an episode is dynamic. This property is advantageous for multi-task learning and adaptation because programs that can switch control code in and out of the execution path on-the-fly are able to dynamically integrate general, re-useable code for related tasks \textit{and} specialized code for disjoint tasks. We demonstrate in Section \ref{laikago_sec} how this improves performance for the quadruped task. 
Each CADF  receives 4 scalars, 2 vectors, and 2 indices as input, and execution of the function is conditional on a $<$ comparison of the first 2 scalars (a configuration chosen for simplicity). The set of operations available is identical to \textcode{GetAction()} except that CADFs may not call each other to avoid infinite recursion. Each CADF uses its own local memory of the same size and dimensionality as the main memory used by \textcode{Setup()} and \textcode{GetAction()}. Their memory is initialized to zero at the start of each episode and is persistent across timesteps, allowing functions to integrate variables over time. Post-execution, the CADF returns the single most recently written index, scalar, and vector from its local memory.

The policy-environment interface and evaluation procedure are illustrated in Fig. \ref{algorithm_evaluation_fig}. Sections \ref{laikago_leg_breaking_sec} and \ref{cartpole_results_sec} provide examples of evolved programs in this representation for the quadruped robot and Cataclysmic Cartpole task, respectively.

\begin{figure}[!h]
            \begin{code}{0.48\textwidth}{4.0in}{codebackground}  
                \codeline{\codecomment{StartEpisode = initialization code.}}
                \codeline{\codecomment{GetAction = control algorithm.}}
                \codeline{\codecomment{Sim = simulation environment.}}
                \codeline{\codecomment{episodes = number of evaluation episodes.}}
                \codeline{\codecomment{sX/vX/mX/iX: scalar/vector/matrix/index memory}}
                \codeline{\codecomment{at address X.}}
                \codeline{\codedef{def} EvaluateFitness(StartEpisode, GetAction):}
                \codeline{\codetab sum\_reward = 0} 
                \codeline{\codetab for e in episodes:}
                \codeline{\codetab \codetab reward = 0}
                \codeline{\codetab \codetab steps = 0}
                \codeline{\codetab \codetab \codecomment{Initialize sX/vX/mX with evolved parameters.}}
                \codeline{\codetab \codetab \codecomment{iX is initialized to zero.}}
                \codeline{\codetab \codetab StartEpisode()}
                \codeline{\codetab \codetab \codecomment{Set environment initial conditions.}}
                \codeline{\codetab \codetab state = Sim.Reset()}
                \codeline{\codetab \codetab while (!Sim.Terminal()):}
                \codeline{\codetab \codetab \codetab \codecomment{Copy state to memory, will be accessible}} 
                \codeline{\codetab \codetab \codetab \codecomment{to GetAction.}}
                \codeline{\codetab \codetab \codetab v1 = state}
                \codeline{\codetab \codetab \codetab \codecomment{Execute action-prediction instructions.}}
                \codeline{\codetab \codetab \codetab GetAction(state)} \label{eval_algo_get_action_line}
                \codeline{\codetab \codetab \codetab if Sim.NumAction() > 1:}
                \codeline{\codetab \codetab \codetab \codetab action = v4} 
                \codeline{\codetab \codetab \codetab else:}
                \codeline{\codetab \codetab \codetab \codetab action = s3} 
                \codeline{\codetab \codetab \codetab state = Sim.Update(action)}
                \codeline{\codetab \codetab \codetab reward += Reward(state, action)}
                \codeline{\codetab \codetab \codetab steps += 1}
                \codeline{\codetab \codetab sum\_reward += reward}
                \codeline{\codetab \codetab sum\_steps += steps}
                \codeline{\codetab \codereturn{return} sum\_reward/episodes, sum\_steps/episodes}
            \end{code}
\caption{Evaluation process for an evolved control algorithm. The single-objective evolutionary search uses the mean episodic reward as the algorithm's \textit{fitness}, while the multi-objective search optimizes two fitness metrics: mean reward (first return value) and mean steps per episode (second return value).}
\label{algorithm_evaluation_fig}
\end{figure}

\subsection{Evolutionary Search}\label{evo_search_sec}
Two evolutionary algorithms are employed in this work: Multi-objective search with the Nondominated Sorting genetic algorithm II (NSGA-II) \cite{deb2002} and single-objective search with Regularized evolution (RegEvo) \cite{real2019regevo, real2020automl}. Both search algorithms iteratively update a population of candidate control programs using an algorithmic model of the Darwinian principle of natural selection. The generic steps for evolutionary search are:
\begin{enumerate}
    \item Initialize a population of random control programs.
    \item Evaluate each program in the task (Fig. \ref{algorithm_evaluation_fig}).\label{GA_evaluate_step}
    \item Select promising programs using a task-specific fitness metric (See Fig. \ref{algorithm_evaluation_fig} caption).
    \item Modify selected individuals through crossover and then mutation (Fig. \ref{code_mutation_algs}). \label{GA_variation_step}
    \item Insert new programs into the population, replacing some proportion of existing individuals.
    \item Go to step \ref{GA_evaluate_step}.
\end{enumerate}
For the purposes of this study, the most significant difference between NSGA-II and RegEvo is their selection method. NSGA-II identifies promising individuals using multiple fitness metrics (e.g., forward motion \textit{and} stability) while RegEvo selects based on a single metric (forward motion). Both search methods simultaneously evolve: (1) Initial algorithm parameters (\ie initial values in floating-point memory \textcode{sX}, \textcode{vX}, \textcode{mX}), which are set by \textcode{StartEpisode()}; and (2) Program content of the \textcode{GetAction()} function and CADFs. 

\subsubsection{Multi-Objective Search}
In the Quadruped robot tasks, the goal is to build a controller that continuously walks at a desired pace in the presence of motor malfunctions. It is critical that real-world robots avoid damage associated with falling, and the simplest way for a robot to achieve this is by standing relatively still and not attempting to move forward after it detects damage. As such, this domain is well suited to multi-objective search because walking in the presence of unpredictable dynamics while maintaining stability are conflicting objectives that must be optimized simultaneously. In this work, we show how NSGA-II maintains a diverse population of control algorithms covering a spectrum of trade-offs between forward motion and stability. From this diverse population of partial solutions, or \textit{building blocks}, evolutionary search operators (mutation and cross-over) can build policies that are competent in both objectives. NSGA-II objective functions and constraints for the quadruped robot task are discussed in Section \ref{laikago_sec}. 

\subsubsection{Single-Objective Search}
The Cataclysmic Cartpole task provides a challenging adaptation benchmark environment without the safety constraints and simulation overhead of the real-world robotics task. To further simplify our study of adaptation and reduce experiment time in this task, we adopt the RegEvo search algorithm and optimize it for fast experimentation. Unlike NSGA-II, asynchronous parallel workers in RegEvo also perform selection, which eliminates the bottleneck of waiting for the entire population to be evaluated prior to ranking, selecting, and modifying individuals.  

\textit{Crossover and Mutation Operators:}
We use a simple crossover operator that swaps a randomly selected CADF between two parent algorithms. Since all CADFs have the same argument list and return value format, no signature matching is required to select crossover points. If either parent algorithm contains no CADFs, one randomly selected parent is returned. Post-crossover, the child program is subject to stochastic mutation, which adds, removes, or modifies code using operators listed in Table \ref{mutation_op_tbl}. 

\subsection{Algorithm Configurations and Baselines}\label{temporal_memory_sec}
Temporal memory is the primary mental system that allows an organism to change, learn, or adapt during its lifetime. In order to predict the best action for a given situation in a dynamic environment, the policy must be able to compare the current situation with past situations and actions. This is because generating an appropriate action depends on the current state \textit{and} a prediction of how the environment is changing. Our evolved algorithms are able to adapt partly because they are \textbf{stateful}: the contents of their memory (\textcode{sX}, \textcode{vX}, \textcode{mX}, and \textcode{iX}) are persistent across timesteps of an episode. 

We compare ARZ against stateless and stateful baselines. These policy architectures consist, respectively, of multilayer perceptrons (\textbf{MLP}) and long short-term memory (\textbf{LSTM}) networks whose parameters to be optimized are purely continuous. Therefore, we use Augmented Random Search (ARS) \cite{NEURIPS2018_7634ea65}, which is a state-of-the-art continuous optimizer and has been shown to be particularly effective in learning robot locomotion tasks \cite{yu19, lee2022pi}. In comparison, Proximal Policy Optimization \cite{ppo} underperformed significantly; we omit the results and leave investigation for future work. All methods were allowed to train until convergence with details in Supplement \ref{baseline_supp}.

\section{Non-Stationary Task Domains}\label{laikago_sec}

We consider two different environments: a realistic simulator for a quadruped robot and the novel Cataclysmic Cartpole. In both cases, policies must handle changes in the environment's transition function that would normally impede their proper function. These changes might be sudden or gradual, and no sensor input is provided to indicate when a change is occurring or how the environment is changing.

\subsection{Quadruped Robot} \label{laikago_setup_sec}

We use the Tiny Differentiable Simulator \cite{Heiden-2021-1140} to simulate the Unitree Laikago robot \cite{laikagounitree}. It is a quadruped robot with 3 actuated degrees of freedom per leg. Thus the action space has 12-dimensional real values corresponding to desired motor angles. A Proportional-Derivative controller is used to track these desired angles. The observation space includes 37 real values describing the angle and velocity for each joint as well as the position, orientation, and velocity of the robot body. Each episode begins with the robot in a stable upright position and continues for a maximum of 1000 timesteps (10 seconds). Each action suggested by the policy is repeated for 10 consecutive steps. 

The goal of the non-stationary quadruped task is to move forward (x-axis) at 1.0 meters/second. Adaptation must handle sudden \textit{leg-breaking} in which all joints on a single, randomly selected leg suddenly become passive at a random time within each episode. The leg effectively becomes a double pendulum for the remainder of the episode. The episode will terminate early if the robot falls and this results in less return. We design the following reward function:
\begin{equation}
    r(t) = 1.0-2*|v(t)-\bar{v}|-||\vec{a}(t) - \vec{a}(t-1)||_2, \label{laikago_fitness}
\end{equation}
where the first term 1.0 is the survival bonus, $\bar{v}$ is the target forward velocity of 1 m/s, $v(t)$ is the robot's current forward velocity, and $\vec{a}(t)$ and $\vec{a}(t-1)$ are the policy's current and previous action vectors. This reward function is \textit{shaped} to encourage the robot to walk at a constant speed for as long as possible while alleviating motor stress by minimizing the change in the joint acceleration. In the context of multi-objective search, maximizing the mean of Equation \ref{laikago_fitness} over a maximum of 1000 timesteps is Objective 1. To discourage behaviors that deviate too much along the y-axis, we terminate an episode if the robot's y-axis location exceeds $\pm 3.0$ meters. Objective 2 is simply the number of timesteps the robot was able to survive without falling or reaching this y-axis threshold. Importantly, we are not interested in policies that simply stand still. Thus, if Objective 2 is greater than $400$ \textit{and} Objective 1 is less than $50$, both fitnesses are set to 0. As shown in Fig. \ref{nsga2_constraints}, these fitness constraints eliminate policies that would otherwise persist in the population without contributing to progress on the forward motion objective. 

\subsection{Cataclysmic Cartpole Environment}\label{cartpole_sec}

To study the nature of adaptation in more detail, we introduce a new, highly challenging but computationally simple domain called Cataclysmic Cartpole in which multiple aspects of the classic Cartpole (\cite{sutton18}) physics are made dynamic. Adaptation must handle the following non-stationary properties:

\begin{itemize}
    \item Track Angle: The track tilts to a random angle at a random time. Because the robot's frame of reference for the pole angle ($\theta$) is relative to the cart, it must figure out the new direction of gravity and \emph{desired} value of $\theta$ to maintain balance, and respond quickly enough to keep the pole balanced. The track angle is variable in [-15, 15] degrees. This simulates a change in the external environment.
    \item Force: A force multiplier $f$ is applied to the policy's action such that its actuator strength may increase or decrease over time. The policy's effective action is $f \times \textrm{action}$, where $f$ changes over time within the range [0.5, 2]. This simulates a drop in actuator strength due to a low battery, for example.
    \item Damping: A damping factor $D$ simulates variable joint friction by modifying joint torque as $\tau_D = -D\dot{q_r}$, where $\dot{q_r}$ is the joint velocity (see eqns. 2.81, 2.83 in \cite{tdssimulatornotes}). This simulates joint wear and tear.
    $D$ changes over time in the range [0.0, 0.15].
 
\end{itemize}

Each type of change is controlled by a single parameter. We investigate two schedules for how these parameters might change during an episode, illustrated in Fig. \ref{change_schedule_fig}.
\section{Results}
\label{results_sec}
 
\subsection{Quadruped Leg-Breaking}\label{laikago_leg_breaking_sec}

\subsubsection{Comparison with Baselines} 
\begin{figure}[!h]
     \centering
     \begin{subfigure}[b]{0.23\textwidth}
         \centering
         \includegraphics[height=4cm]{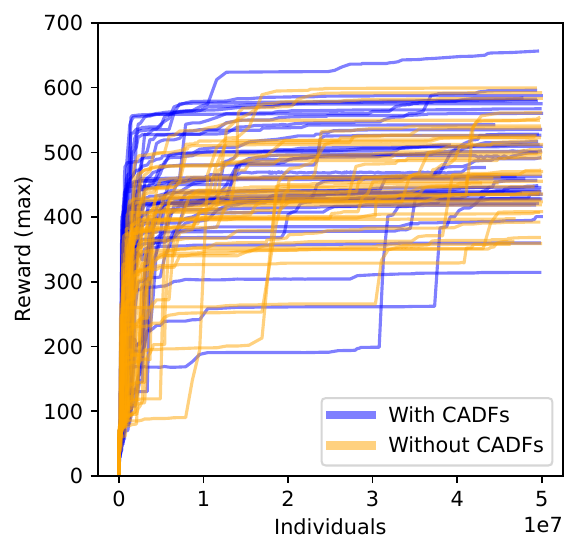}
         \caption{Evolution progress}
         \label{gp_break_train_o1_fig}
     \end{subfigure}
     \begin{subfigure}[b]{0.22\textwidth}
         \centering
         \includegraphics[height=4cm]{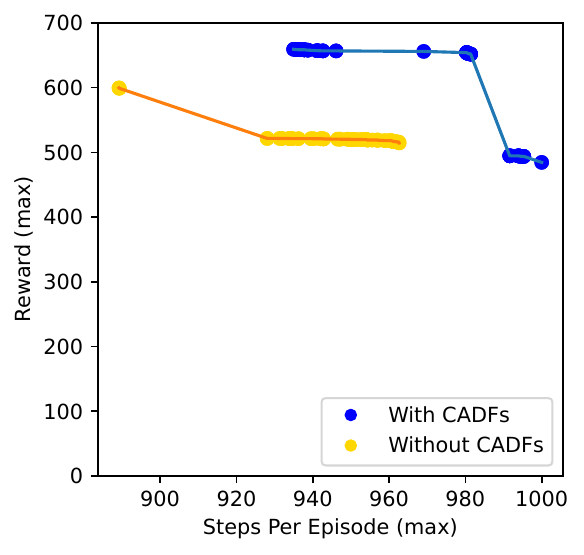}
         \caption{Best Pareto fronts}
         \label{gp_break_best_pareto_fig}
     \end{subfigure}
        \caption{CADFs speed up evolution on average and produced the best final result. (a) shows ARZ search data recorded over 20 independent repeats with and without the use of CADFs. The horizontal axis for (a) shows the total number of individual programs evaluated, while the vertical axis shows mean return (Equation \ref{laikago_fitness}) over 32 episodes for the single best individual discovered so far. (b) shows Pareto fronts for the single repeats with max reward from each experiment. Each point in (b) represents the bi-objective fitness of one control program.}
        \label{laikago_train_curves_fig}
\end{figure}

ARZ---with the inclusion of CADFs---is the only method that produced a viable control policy in the leg-breaking task. This problem is exceedingly difficult: finding a policy that maintains smooth locomotion and is robust to leg breaking requires 20 evolution experiment repetitions (Fitness $>600$ in Fig. \ref{gp_break_train_o1_fig}). In Fig. \ref{gp_break_train_o1_fig}, training fitness between 500 and 600 typically indicates either (1) a viable forward gait behavior that is only robust to 3/4 legs breaking or (2) a policy robust to \emph{any} leg breaking but which operates at a high frequency not viable for a real robot, with its reward being significantly penalized by fitness shaping as a result. Within the single best repeat, the NSGA-II search algorithm produces a variety of policies with performance trade-offs between smooth forward locomotion (reward objective) and stability (steps objective), Fig. \ref{gp_break_best_pareto_fig}. From this final set of individuals, we select a single policy to compare with the single best policy from each baseline. Due to practical wall-clock time limits, we were only able to train both ARS+MLP and ARS+LSTM policies up to $10^{6}$ trials in total, but found that under this sample limit, even the best ARS policy only achieved a reward of 360, much lower than the 570 found by the best ARZ policy, suggesting that ARZ can even be more sample efficient than standard neural network baselines.

\begin{figure}[!h]
	\centering
	\includegraphics[width=1.0\columnwidth]{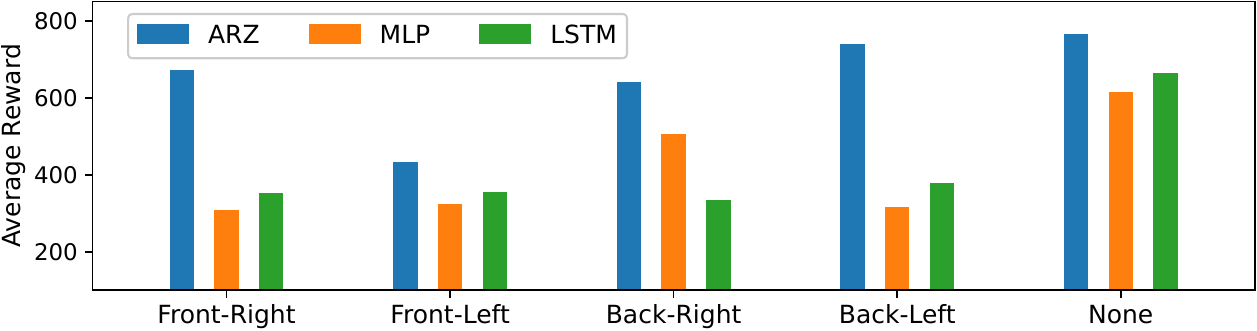}
	\caption{ARZ discovers the only policy that can adapt to \text{any} leg breaking. The plot shows test results for the single best policy from ARZ and ARS baselines (MLP and LSTM) in the \emph{mid-episode leg-breaking} task. For each leg, bars show mean reward over 100 episodes in which that leg is broken at a randomly selected timestep. A reward $<400$ in any column indicates the majority of test episodes for that leg ended with a fall.}
	\label{leg_breaking_test_fig}
\end{figure}

Fig. \ref{leg_breaking_test_fig} confirms that ARZ is the only method capable of building a controller that is robust to multiple different legs breaking mid-episode. We plot post-training test results for one champion ARZ policy in comparison with the single-best controller discovered by ARS+MLP and ARS+LSTM.  ARZ's adaption quality (as measured by mean reward) is superior to baselines in the case of each individual leg, and its performance on the stationary task (See "None" in Fig. \ref{leg_breaking_test_fig}) is significantly better than any other method. Interestingly, Fig. \ref{leg_breaking_test_fig} indicates that the MLP also learned a policy that is robust to the specific case of the back-right leg breaking. Unlike ARZ, it is unable to generalize this adaptation to any other leg. Finally, while the LSTM policy performed better than the MLP on the stationary task, it fails to adapt to any of the leg-breaking scenarios.

\begin{figure}[!h]
     \centering
     \begin{subfigure}[b]{0.155\textwidth}
         \centering
         \includegraphics[width=\textwidth]{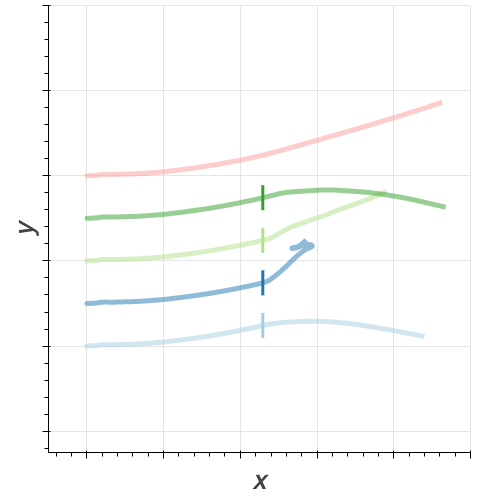}
         \caption{ARZ}
         \label{gp_break_traj_fig}
     \end{subfigure}
     \begin{subfigure}[b]{0.155\textwidth}
         \centering
         \includegraphics[width=\textwidth]{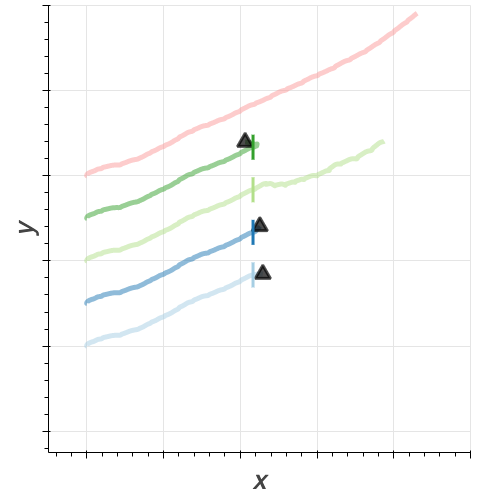}
         \caption{MLP}
         \label{mlp_break_traj_fig}
     \end{subfigure}
     \begin{subfigure}[b]{0.155\textwidth}
         \centering
         \includegraphics[width=\textwidth]{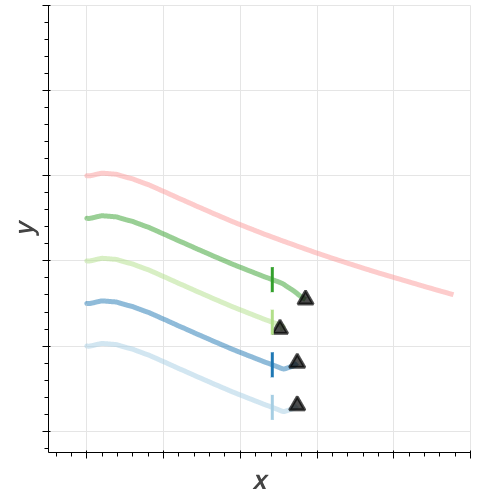}
         \caption{LSTM}
         \label{lstm_break_traj_fig}
     \end{subfigure}
        \caption{ARZ discovers the only policy that consistently avoids falling. Plot shows sample trajectories in each leg-breaking task. The vertical bar indicates the change point (step 500). $\boldsymbol{\blacktriangle}$ indicates that the robot fell over. Each plot shows 4 test episodes in which a unique leg breaks. From top to bottom, the affected legs are: None, Back-Left, Back-Right, Front-Left, Front-Right.}
        \label{laikago_traj_fig}
\end{figure}

Visualizing trajectories for a sample of 5 test episodes from Fig. \ref{leg_breaking_test_fig} confirms that the ARZ policy is the only controller that can avoid falling in all scenarios, although in the case of the front-left leg breaking, it has trouble maintaining forward motion, Fig. \ref{laikago_traj_fig}. This is reflected in its relatively weak test reward for the front-left leg (See Fig. \ref{leg_breaking_test_fig}). The MLP policy manages to keep walking with a broken back-right leg but falls in all other dynamic tasks. The LSTM, finally, is only able to avoid falling in the stationary task in which all legs are reliable. 

\subsubsection{On Simplicity and Interpretability}
The policy for the Quadruped Leg-Breaking task discovered by evolutionary search is presented in Fig. \ref{quadruped_algo}. This algorithm uses 608 parameters and can be expressed in less than 40 lines of code, executing at most 2080 floating point operations (FLOPs) per step. This should be contrasted with the number of parameters and FLOPs expended in the baseline MLP/LSTM models, which use more than 2.5k/9k parameters and 5k/18k FLOPs per step, respectively. A detailed account of how these numbers were obtained can be found in Section \ref{ap:complexity}. We note that each function possesses its own variables and memory, which persists throughout the run. The initialization value for the variables are tuned for the \texttt{GetAction} function, thus counted as parameters, while they are all set to zero for \texttt{f}.

\begin{figure}[!h]
  \centering
  \includegraphics[width=0.95\columnwidth]{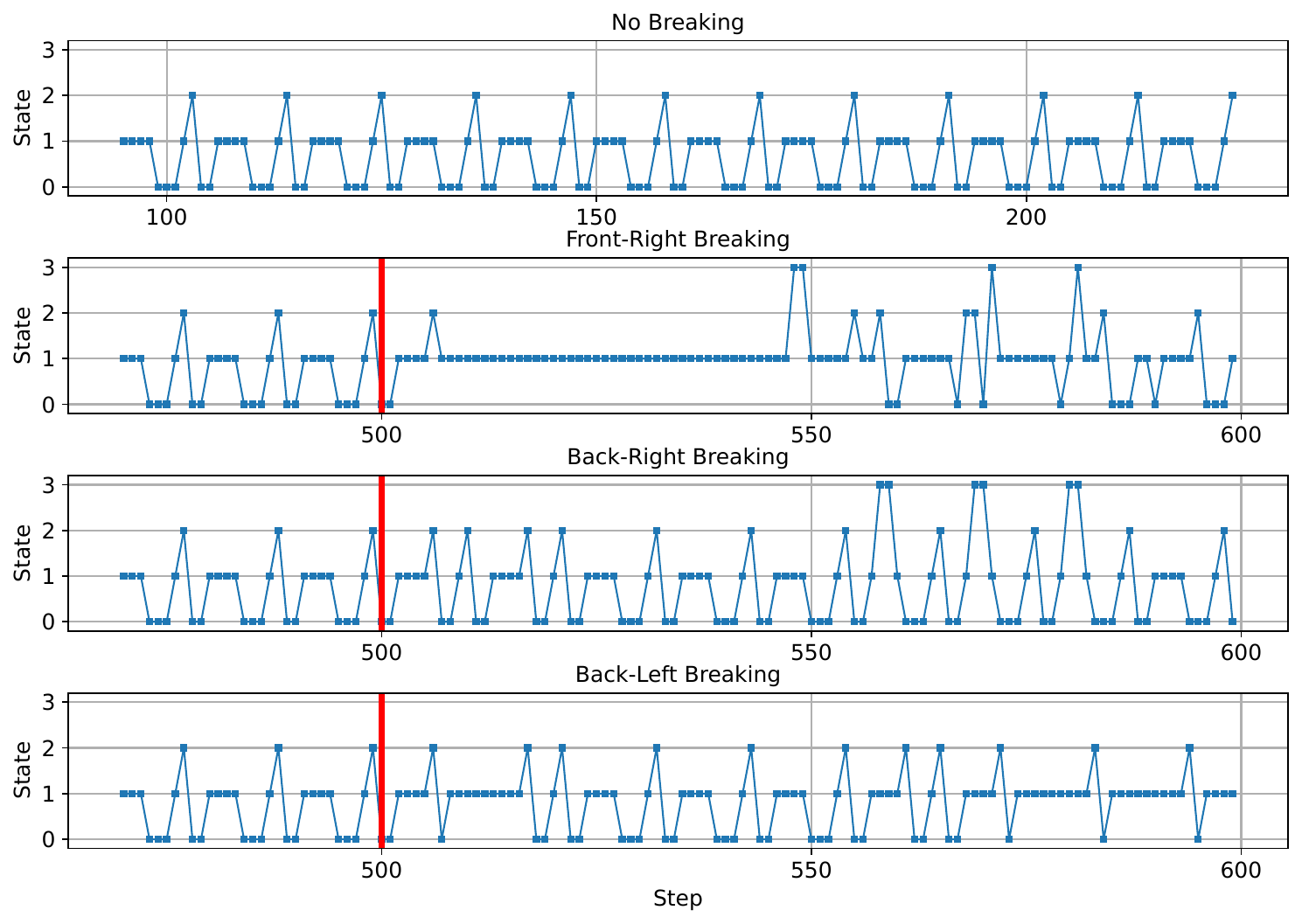}
  \caption{State trajectories of various leg-breaking patterns. The leg-breaking event is marked by a vertical red line. Note that different leg breaking patterns result in different state trajectories. We conjecture that these trajectories serve as signals that trigger the adaptive response in the algorithm.}
  \label{state-trajectories}
\end{figure}

Here we provide an initial analysis of the ARZ policy, leaving a full analysis and interpretation of the algorithm to future work. The key feature of the algorithm is that it discretizes the input into four states, and the action of the quadruped is completely determined by its internal state and the discrete label. The temporal transitions of the discretized states show a stable periodic motion when the leg is not broken, and the leg-breaking introduces a clear disruption in this pattern, as shown in Fig. \ref{state-trajectories}. This being a stateful algorithm with multiple variables accumulating and preserving variables from previous steps, we conjecture that the temporal pattern of the discrete states serves as a signal for the adaptive behavior of the quadruped.

We now expand upon how the continuous input signal is discretized in the ARZ algorithm presented in Fig. \ref{quadruped_algo}. We first observe that the only way the incoming observation vector interacts with the rest of the algorithm is by forming scalar \texttt{s12}, by taking an inner-product with a dynamical vector \texttt{v3} (the second of the three red-colored lines of code). The scalar \texttt{s12} affects the action only through the two \texttt{if} statements colored in red. Thus the effect of the input observation on the action is entirely determined by the relative position of the scalar \texttt{s12} with respect to the two decision boundaries set by the scalars \texttt{s15} and \texttt{s7}. In other words, the external input of the observation to the system is effectively discretized into four states: 0 (\texttt{s12 $\leq$ s15, s7}), 1 (\texttt{s15, s7 < s12}), 2 (\texttt{s7 < s12 $\leq$ s15}) or 3 (\texttt{s15 < s12 $\leq$ s7}).

\begin{figure}[!h]
  \centering
  \includegraphics[width=0.95\columnwidth]{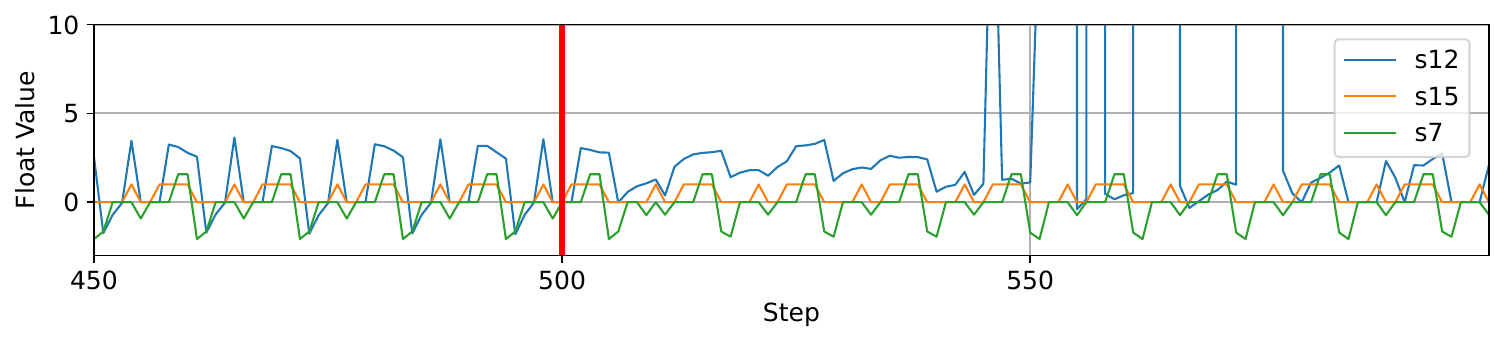}
  \caption{The scalar values \texttt{s12}, \texttt{s15} and \texttt{s7} of the quadruped during front-leg breaking. Note the consistent periodic behavior of the scalars \texttt{s15} and \texttt{s7} despite leg breaking, marked by the vertical red line. The same periodicity is observed for all leg-breaking scenarios analyzed.}
  \label{periodic-scalars}
\end{figure}

Thus external changes in the environment, such as leg breaking, can be accurately detected by the change in the pattern of the state trajectory, because the variables \texttt{s7} and \texttt{s15} defining the decision boundary of the states form a stable periodic function in time. We demonstrate this in Fig. \ref{periodic-scalars}, where we plot the values of the three scalars \texttt{s12}, \texttt{s15} and \texttt{s7} for front-leg breaking, whose occurrence is marked by the vertical red line. Despite the marked change of behavior of the input \texttt{s12} after leg-breaking, we see that the behavior of the two scalars \texttt{s7} and \texttt{s15} are only marginally affected. Intriguingly, the behavior of the scalar registers \texttt{s7} and \texttt{s15} resemble that of central pattern generators in biological circuits responsible for generating rhythmic movements \cite{marder2001central}.

\begin{figure}[!h]
     \centering
     \begin{subfigure}[b]{0.2\textwidth}
         \centering
         \includegraphics[width=\textwidth]{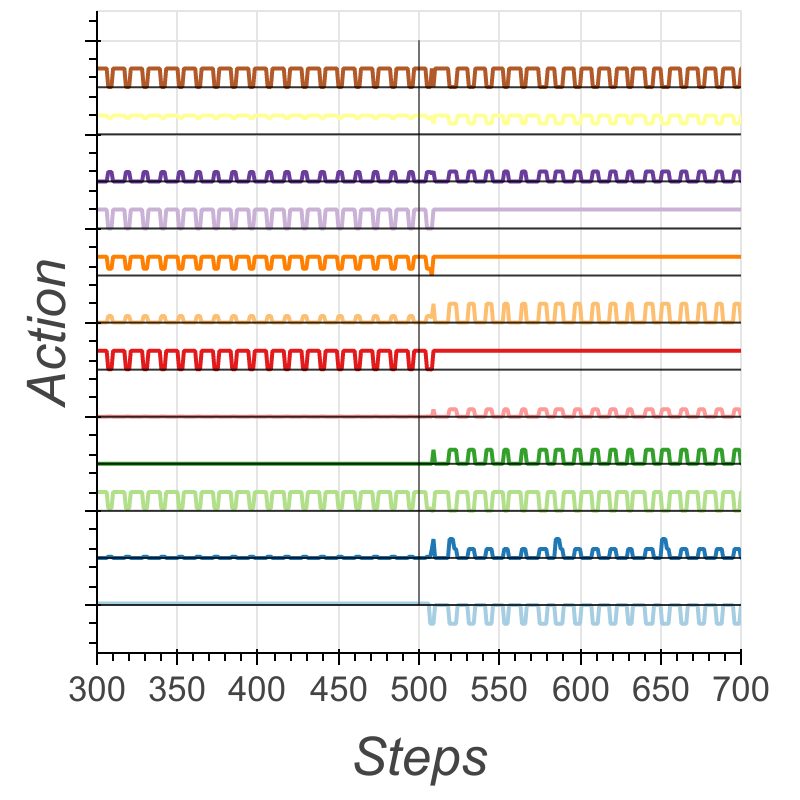}
         \caption{Actions}
         \label{gp_break_acts_trial1_fig}
     \end{subfigure}
     \begin{subfigure}[b]{0.275\textwidth}
         \centering
         \includegraphics[width=\textwidth]{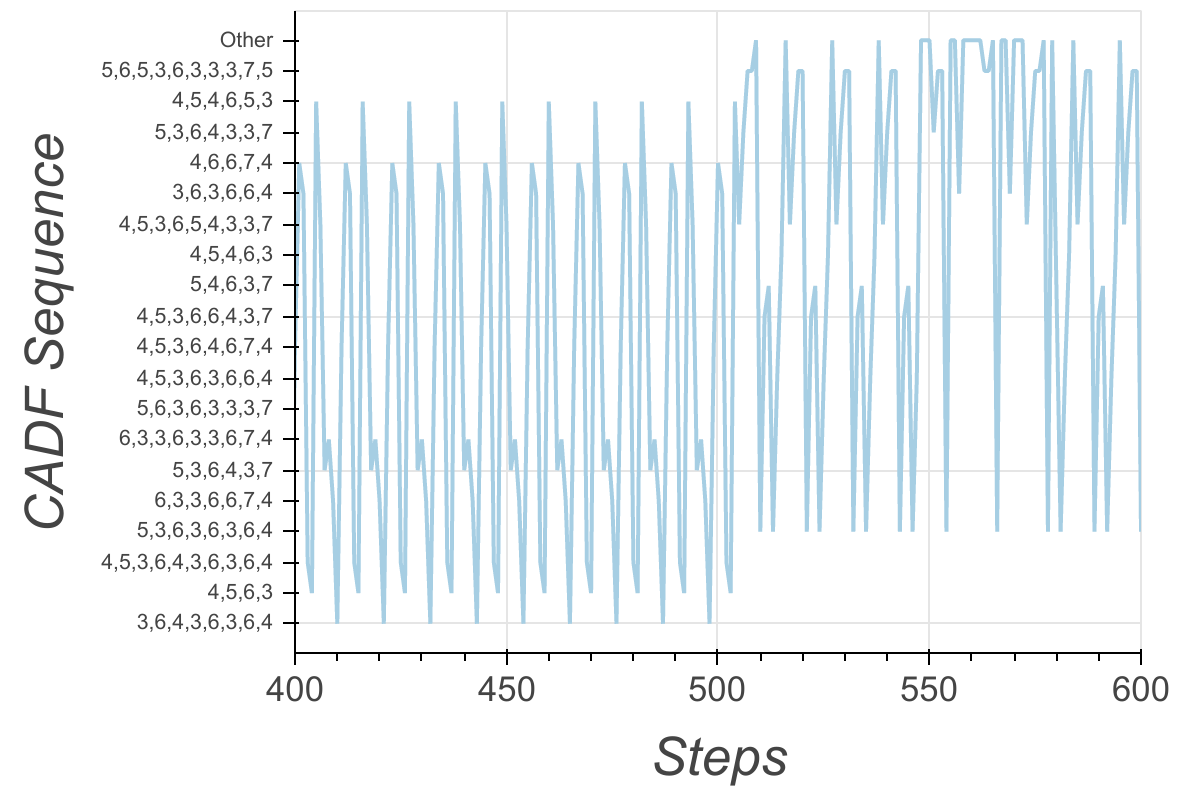}
         \caption{CADF call sequences}
         \label{gp_fcalls_trial1_fig}
     \end{subfigure}
        \caption{ARZ policy behavior changes when Front-Left leg breaks mid-episode (step 500), as shown by the dynamics of the actions and the program control flow due to CADFs.}
        \label{laikago_actions_fig}
\end{figure}

The policy's ability to quickly identify and adapt to multiple unique failure conditions is clear in Fig. \ref{gp_break_acts_trial1_fig}, which plots the controller's actions one second before and after a leg breaks. We see a clear, instantaneous change in behavior when a leg fails. This policy is able to identify when a change has occurred and rapidly adapt. Fig. \ref{gp_fcalls_trial1_fig} shows the particular sequence of CADFs executed at each timestep before and after the change, indicating that CADFs do play a role in the policy's ability to rapidly adjust its behavior. Indeed, only evolutionary runs that included CADFs were able to discover a policy robust to any leg breaking.

\subsection{Cataclysmic Cartpole}\label{cartpole_results_sec}

Introducing a novel benchmark adaptation task is an informative addition to results in the realistic quadruped simulator because we can empirically adjust the nature of the benchmark dynamics until they are significant enough to create an \textit{adaptation gap}: when stateless policies (i.e., MLP generalists) fail to perform well because they cannot adapt their control policy in the non-stationary environment (See Section \ref{ap:cartpole_experiments} for details.). Having confirmed that Cataclysmic Cartpole requires adaptation, we only examine stateful policies in this task.

\subsubsection{Comparison with Baselines} 
\begin{figure}[!h]
	\centering
	\includegraphics[width=1.0\columnwidth]{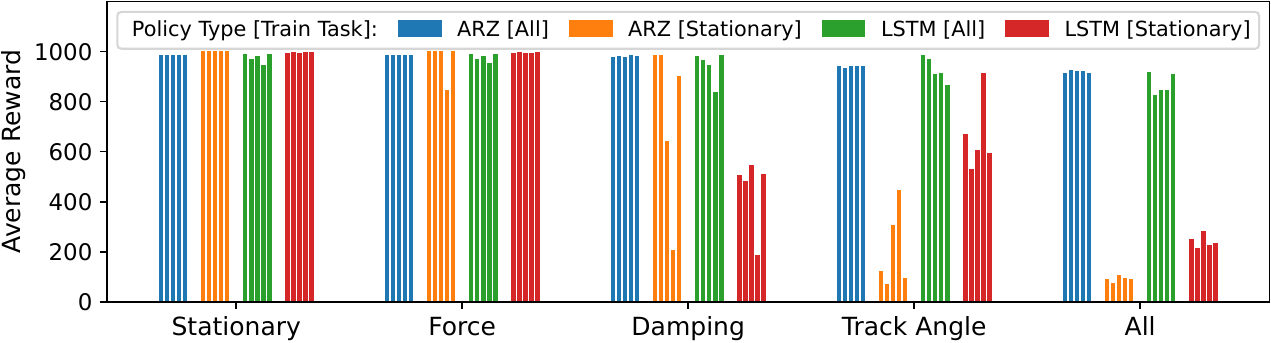}
	\caption{Post-evolution test results in the Cataclysmic Cartpole \emph{continuous-change}  task. Legend indicates policy type and search task. \textcode{[All]} marks policies exposed to all tasks during evolution. ARZ and LSTM both solve this adaptation task, and no direct transfer from stationary tasks to dynamic tasks is observed. The best 5 policies from each experiment are shown.}
	\label{cartpole_metaval_continous_a_fig}
\end{figure}

\begin{figure}[!h]
	\centering
	\includegraphics[width=1.0\columnwidth]{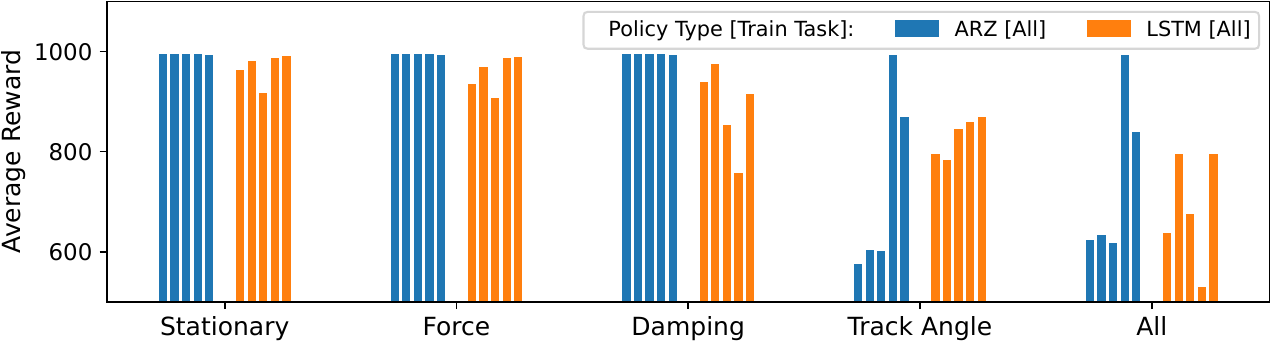}
	\caption{Post-evolution test results in the Cataclysmic Cartpole \emph{sudden-change}  task. \textcode{[All]} marks policies exposed to all tasks during evolution. ARZ discovers the only policy that adapts to all sudden-change Cataclysmic Cartpole tasks. The best 5 policies from each experiment are shown.}
	\label{cartpole_metaval_sudden_a_fig}
\end{figure}

In Cataclysmic Cartpole, we confirm that ARZ produces superior control relative to the (stateful) ARS+LSTM baseline in tasks with a sudden, dramatic change. Fig. \ref{cartpole_metaval_continous_a_fig} and \ref{cartpole_metaval_sudden_a_fig} show testing that was done after the search is complete. A fitness score of 800 indicates the policy managed to balance the pole for $\approx 800$ timesteps, surviving up to the last point in an episode with any active dynamics (See Fig. \ref{change_schedule_fig}). "Stationary" is the standard Cartpole task while "Force", "Damping", and "Track Angle" refer to Cartpole with sudden or continuous change in these parameters \textit{only} (See Section \ref{cartpole_sec}). "All" is the case where all change parameters are potentially changing simultaneously. Legends indicate the policy type and corresponding task type used during evolution. First, note that strong adaptable policies do not emerge from ARZ or ARS+LSTM evolved in the stationary task alone (See ARZ [Stationary] and LSTM [Stationary]), implying that proficiency in the stationary task does not directly transfer to any non-stationary configuration. However, when exposed to non-stationary properties during the search, ARZ and ARS+LSTM discover policies that adapt to all sudden and continuous non-stationary tasks. ARZ is significantly more proficient in the sudden change tasks (Fig. \ref{cartpole_metaval_sudden_a_fig}), achieving near perfect scores of $\approx 1000$ in all tasks. In continuous change, the single best LSTM policy achieves the best multitasking performance with a stronger score than ARZ on the Track Angle problem, and it is at least as proficient as ARZ on all other tasks. However, unlike the LSTM network, ARZ policies are uniquely interpretable. 

\subsubsection{On Simplicity and Interpretability} 
Here we decompose an ARZ policy to provide a detailed explanation of how it integrates state observations over time to compute optimal actions in a changing environment. An example of an algorithm discovered in the ARZ [All] setting of Fig. \ref{cartpole_metaval_continous_a_fig} is presented in Fig. \ref{sample_stateful_all_algo}. Note that CADFs were not required to solve this task and have thus been omitted from the search space in order to simplify program analysis. What we find are three accumulators that collect the history of observation and action values from which the current action can be inferred.

\begin{figure}[!h]
    \begin{code}{0.48\textwidth}{1.15in}{codebackground}
        \scriptsize
        \codeline{\codecomment{sX: scalar memory at address X.}}
        \codeline{\codecomment{obs: vector [x, theta, x\_dot, theta\_dot].}}
        \codeline{\codecomment{a, b, c: fixed scalar parameters.}}
        \codeline{\codecomment{V, W: 4-dimensional vector parameters.}}
        \codeline{\codedef{def} GetAction(obs, action):}
        \codeline{\codetab s0 = a * s2 + action}
        \codeline{\codetab s1 = s0 + s1 + b * action + dot(V, obs)}
        \codeline{\codetab s2 = s0 + c * s1}
        \codeline{\codetab action = s0 + dot(obs, W)}
        \codeline{\codetab \codereturn{return} action}
    \end{code}
\caption{Sample stateful action function evolved on the task where all parameters are subject to continuous change (ARZ [All] in Fig.\ \ref{cartpole_metaval_continous_a_fig}). Code shown in Python.}
\label{sample_stateful_all_algo}
\end{figure}

This algorithm uses 11 variables and executes 25 FLOPs per step. Meanwhile, the MLP and LSTM counterparts use more than 1k and 4.5k parameters, expending more than 2k and 9k FLOPs per step, respectively. More details for this computation are presented section \ref{ap:complexity}.
 
There are two useful ways to view this algorithm. First, by organizing the values of \textcode{s0}, \textcode{s1}, and \textcode{s2} at step $n$ into a vector $Z_n$, which can be interpreted as a vector in latent space of $d=3$ dimensions, we find that the algorithm can be expressed in the form:
$s_{n+1} = \text{concat}(\text{obs}_{n+1}\,,~\text{act}_n)$; 
$Z_{n+1} = \widetilde{U} \cdot Z_n + \widetilde{P} \cdot s_{n+1}$;
$\text{act}_{n+1} = \widetilde{A}^T \cdot Z_{n+1} + \widetilde{W}^T \cdot s_{n+1}$,
with the projection matrix $\widetilde{P}$ that projects the state vector to the latent space, and a $d \times d$ evolution matrix $\widetilde{U}$. This is a linear recurrent neural network with internal state $Z_n$. The second way to view the algorithm is to interpret it as a generalization of a proportional–integral–derivative (PID) controller. This can be done by first explicitly solving the recurrent equations presented above and taking the continuous limit. Introducing a single five-dimensional state vector $s(t) = [x(t), \theta(t), \dot{x}(t), \dot{\theta}(t), \text{act}(t)]$, and $d$-dimensional vectors $u$, $v$, and $w$, a five-dimensional vector $p$ and a constant term $c$, the algorithm in the continuous time limit can be written in the form: 
$\text{act}(t) = c + w^T \cdot U^t \cdot u + p^T \cdot s(t) + v^T \cdot \int_0^t d\tau ~ U^{t-\tau} \cdot P \cdot s(\tau)$
where $P$ and $U$ are the continuous-time versions of $\widetilde{P}$ and $\widetilde{U}$. In our particular discovered algorithm (Fig. \ref{sample_stateful_all_algo}), $d$ happens to be 3. Notice that the integration measure now has a time-dependent weight factor in the integrand versus the conventional PID controller. Further derivations, discussions, and interpretations regarding this algorithm are presented in the supplementary material.
\section{Conclusion and Discussion}
\label{discussion_sec}
We have shown that using ARZ to search simultaneously in \textit{program space} and \textit{parameter space} produces proficient, simple, and interpretable control algorithms that can perform zero-shot adaptation, rapidly changing their behavior to maintain near-optimal control in environments that undergo radical change. In the remainder of this section, we briefly motivate and speculate about future work. 

\vspace{2mm} \noindent \textbf{CADFs and the Distraction Dilemma.} In the quadruped robot domain, we have observed that including Conditionally invoked Automatically Defined Functions (CADFs) in our search space improves the expressiveness of evolved control algorithms. In the single best policy, CADFs have been used to discretize the observation space into four states. The action is then completely determined by the internal state of the system and this discretized observation. One interpretation is that this discretization helps the policy define a switching behavior that can overcome the \textit{distraction dilemma}: the challenge for a multi-task policy to balance the reward of excelling at multiple different tasks against the ultimate goal of achieving generalization \cite{hasselt2019}. By contrast, searching only in the parameter space of a hand-designed MLP or LSTM network did not produce policies that can adapt to more than one unique change event (i.e., a single leg breaking). A deeper study of modular/hierarchical policies and their impact on the distraction dilemma is left to future work. 

\vspace{2mm} \noindent \textbf{The Cataclysmic Cartpole Task.} Given the computationally intensive nature of simulating a real robot, we felt compelled to also include a more manageable toy task where adaptation matters. This led to the \emph{Cataclysmic Cartpole} task. We found it useful for doing quick experiments and emphasizing the power and interpretability of ARZ results. We hope that it may also provide an easily reproducible environment for use in further research.

\vspace{2mm} \noindent \textbf{Adapting to Unseen Task Dynamics.} Looking to the future, we have included detailed supplementary material which raises an open and ambitious question: how can we build adaptive control policies without any prior knowledge about what type of environmental change may occur in the future? Surprisingly, preliminary results with ARZ on the cataclysmic cartpole task suggest that injecting partial-observability and dynamic actuator noise during evolution (training) can act as a general surrogate for non-stationary task dynamics \ref{ap:cartpole_experiments}. In preliminary work, we found this to support the emergence of policies that can adapt to novel task dynamics that were \textit{not} experienced during search (evolution). This was not possible for our LSTM baselines. If true, this would be significant because it implies we might be able to evolve proficient control policies without complete prior knowledge of their task environment dynamics, thus relaxing the need for an accurate physics simulator. Future work may investigate the robustness of this preliminary finding.

\section*{Author Contributions}
\footnotesize
SK and ER led the project. ER and JT conceived the project and acted as principal advisors. All authors contributed to the methodology. 
SK, MM, PN, and DP ran the evolution experiments. XS ran the baselines. MM and DP analysed the algorithms. SK, DP, and MM wrote the paper. All authors edited the paper.

\section*{Acknowledgements}
\footnotesize
We would like to thank Wenhao Yu, Chen Liang, Sehoon Ha, James Lee and the Google Brain Evolution and AutoML groups for technical discussions; Erwin Coumans for physics simulations advice; Erwin Coumans, Kevin Yap, Jacob Budzis, Heng Li, Kaiyuan Wang, and Ryan Gillard for code contributions; and Quoc V. Le, Vincent Vanhoucke, Ed Chi, and Erik Goodman for guidance and support. 



\bibliography{short}

\begin{thebibliography}{10}
\providecommand{\url}[1]{#1}
\csname url@rmstyle\endcsname
\providecommand{\newblock}{\relax}
\providecommand{\bibinfo}[2]{#2}
\providecommand\BIBentrySTDinterwordspacing{\spaceskip=0pt\relax}
\providecommand\BIBentryALTinterwordstretchfactor{4}
\providecommand\BIBentryALTinterwordspacing{\spaceskip=\fontdimen2\font plus
\BIBentryALTinterwordstretchfactor\fontdimen3\font minus
  \fontdimen4\font\relax}
\providecommand\BIBforeignlanguage[2]{{%
\expandafter\ifx\csname l@#1\endcsname\relax
\typeout{** WARNING: IEEEtran.bst: No hyphenation pattern has been}%
\typeout{** loaded for the language `#1'. Using the pattern for}%
\typeout{** the default language instead.}%
\else
\language=\csname l@#1\endcsname
\fi
#2}}

\bibitem{hasselt2019}
M.~Hessel, H.~Soyer, L.~Espeholt, W.~Czarnecki, S.~Schmitt, and H.~van Hasselt,
  ``{Multi-Task Deep Reinforcement [...]]}.''\hskip 1em plus 0.5em minus
  0.4em\relax AAAI, 2019.

\bibitem{kelly21}
S.~Kelly, T.~Voegerl, W.~Banzhaf, and C.~Gondro, ``Evolving hierarchical
  memory-prediction [...]],'' \emph{Genet.\ Program.\ Evolvable Mach.}, 2021.

\bibitem{real2020automl}
E.~Real, C.~Liang, D.~R. So, and Q.~V. Le, ``{AutoML-Zero: Evolving Machine
  Learning Algorithms From Scratch},'' \emph{ICML}, 2020.

\bibitem{laikagounitree}
\BIBentryALTinterwordspacing
``Unitree {Robotics}.'' [Online]. Available: \url{http://www.unitree.cc/}
\BIBentrySTDinterwordspacing

\bibitem{koza1990}
J.~R. Koza and M.~A. Keane, ``Genetic breeding of non-linear optimal control
  strategies [...]],'' in \emph{Analysis and Optimization of Systems}, 1990.

\bibitem{dhebar2022}
Y.~Dhebar, K.~Deb, S.~Nageshrao, L.~Zhu, and D.~Filev, ``{Toward
  Interpretable-AI Policies [...]]},'' \emph{IEEE.\ Trans.\ Cybern.}, 2022.

\bibitem{yu17}
W.~Yu, J.~Tan, C.~K. Liu, and G.~Turk, ``Preparing for the unknown: Learning a
  universal policy [...],'' in \emph{RSS}, 2017.

\bibitem{tobin17}
J.~Tobin, R.~Fong, A.~Ray, J.~Schneider, W.~Zaremba, and P.~Abbeel, ``Domain
  randomization for transferring [...]],'' \emph{CoRR}, 2017.

\bibitem{cully15}
A.~Cully, J.~Clune, D.~Tarapore, and J.-B. Mouret, ``Robots that can adapt like
  animals,'' \emph{Nature}, 2015.

\bibitem{xingyou20}
X.~Song, Y.~Yang, K.~Choromanski, K.~Caluwaerts, W.~Gao, C.~Finn, and J.~Tan,
  ``{Rapidly Adaptable Legged Robots [...]},'' in \emph{IROS}, 2020.

\bibitem{xingyou2019}
X.~Song, W.~Gao, Y.~Yang, K.~Choromanski, A.~Pacchiano, and Y.~Tang,
  ``{ES-MAML:} simple hessian-free meta learning,'' in \emph{ICLR}, 2020.

\bibitem{yu19}
W.~Yu, J.~Tan, Y.~Bai, E.~Coumans, and S.~Ha, ``Learning fast adaptation with
  meta strategy optimization,'' \emph{IEEE Robot.\ Autom.\ Lett.}, 2020.

\bibitem{finn17}
C.~Finn, P.~Abbeel, and S.~Levine, ``Model-agnostic meta-learning for fast
  adaptation of deep networks,'' in \emph{ICML}, 2017.

\bibitem{kumar21}
A.~Kumar, Z.~Fu, D.~Pathak, and J.~Malik, ``{RMA:} rapid motor adaptation for
  legged robots,'' \emph{CoRR}, 2021.

\bibitem{najarro20}
E.~Najarro and S.~Risi, ``{Meta-Learning through Hebbian Plasticity in Random
  Networks},'' \emph{CoRR}, 2020.

\bibitem{floreano2000}
D.~Floreano and J.~Urzelai, ``Evolutionary robots with on-line
  self-organization and behavioral fitness,'' \emph{Neural Networks}, 2000.

\bibitem{timothee2021}
T.~Anne, J.~Wilkinson, and Z.~Li, ``Meta-learning for fast adaptive locomotion
  with uncertainties [...]],'' in \emph{IROS}, 2021.

\bibitem{li2020}
A.~Li, C.~Florensa, I.~Clavera, and P.~Abbeel, ``Sub-policy adaptation for
  hierarchical reinforcement learning,'' in \emph{ICLR}, 2020.

\bibitem{wang2021}
J.~X. Wang, ``Meta-learning in natural and artificial intelligence,''
  \emph{Current Opinion in Behavioral Sciences}, 2021.

\bibitem{brameier07}
M.~Brameier and W.~Banzhaf, \emph{Linear Genetic Programming}.\hskip 1em plus
  0.5em minus 0.4em\relax Springer, 2007.

\bibitem{koza1994genetic}
J.~R. Koza, \emph{Genetic Programming II: Automatic Discovery of Reusable
  Programs}.\hskip 1em plus 0.5em minus 0.4em\relax Cambridge, MA, USA: MIT
  Press, 1994.

\bibitem{deb2002}
K.~Deb, A.~Pratap, S.~Agarwal, and T.~Meyarivan, ``{A fast and elitist
  multiobjective genetic [...]},'' \emph{IEEE Trans.\ Evol.\ Comput.}, 2002.

\bibitem{real2019regevo}
E.~Real, A.~Aggarwal, Y.~Huang, and Q.~V. Le, ``Regularized evolution for image
  classifier architecture search,'' \emph{AAAI}, 2019.

\bibitem{NEURIPS2018_7634ea65}
H.~Mania, A.~Guy, and B.~Recht, ``Simple random search of static linear
  policies is competitive [...]],'' in \emph{NeurIPS}, 2018.

\bibitem{lee2022pi}
K.-H. Lee, O.~Nachum, T.~Zhang, S.~Guadarrama, J.~Tan, and W.~Yu, ``{PI-ARS:
  Accelerating Evolution-Learned [...]]},'' in \emph{IROS}, 2022.

\bibitem{ppo}
J.~Schulman, F.~Wolski, P.~Dhariwal, A.~Radford, and O.~Klimov, ``Proximal
  policy optimization algorithms,'' \emph{CoRR}, 2017.

\bibitem{Heiden-2021-1140}
E.~Heiden, D.~Millard, E.~Coumans, Y.~Sheng, and G.~S. Sukhatme, ``{NeuralSim:
  Augmenting Differentiable [...]]},'' in \emph{ICRA}, 2021.

\bibitem{sutton18}
R.~S. Sutton and A.~G. Barto, \emph{Reinforcement Learning: An
  Introduction}.\hskip 1em plus 0.5em minus 0.4em\relax Cambridge, MA, USA: A
  Bradford Book, 2018.

\bibitem{tdssimulatornotes}
\BIBentryALTinterwordspacing
S.~Sueda, ``Analytically differentiable articulated [...]],'' 2021. [Online].
  Available: \url{https://github.com/sueda/redmax/blob/master/notes.pdf}
\BIBentrySTDinterwordspacing

\bibitem{marder2001central}
E.~Marder and D.~Bucher, ``Central pattern generators and the control of
  rhythmic movements,'' \emph{Current Biology}, 2001.

\bibitem{gillard2023unified}
R.~Gillard, S.~Jonany, Y.~Miao, M.~Munn, C.~de~Souza, J.~Dungay, C.~Liang,
  D.~R. So, Q.~V. Le, and E.~Real, ``Unified functional hashing in automatic
  machine learning,'' \emph{arXiv}, 2023.

\bibitem{stanley2002evolving}
K.~O. Stanley and R.~Miikkulainen, ``Evolving neural networks through
  augmenting topologies,'' \emph{Evolutionary Computation}, 2002.

\bibitem{bergstra2012random}
J.~Bergstra and Y.~Bengio, ``Random search for hyper-parameter optimization,''
  \emph{JMLR}, 2012.

\bibitem{tf_agents}
D.~Hafner, J.~Davidson, and V.~Vanhoucke, ``Tensorflow agents: Efficient
  batched reinforcement learning in tensorflow,'' \emph{CoRR}, 2017.

\bibitem{brockman2016gym}
G.~Brockman, V.~Cheung, L.~Pettersson, J.~Schneider, J.~Schulman, J.~Tang, and
  W.~Zaremba, ``{OpenAI Gym},'' 2016.

\end{thebibliography}
\bibliographystyle{IEEEtran}


\nocite{gillard2023unified}
\nocite{stanley2002evolving,bergstra2012random}
\nocite{NEURIPS2018_7634ea65}
\nocite{ppo}
\nocite{tf_agents}
\nocite{sutton18, brockman2016gym}
\nocite{koza1990,brameier07}
\nocite{real2020automl}

\supplheader{}

\supplsection{Methods Additional Details}{ap:methods}{

\begin{figure}[!h]
   \centering
   \includegraphics[width=0.48\textwidth]{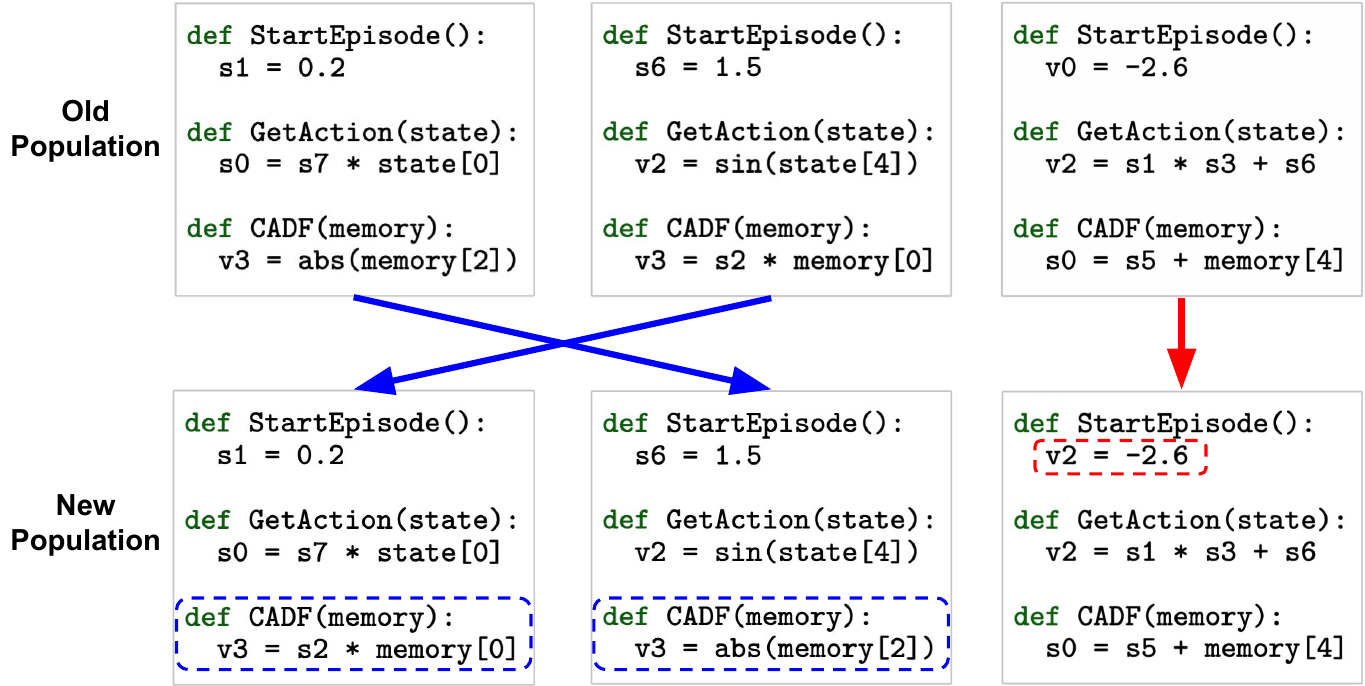}
   \caption{Simplified example of a population of algorithms, modified via \textcolor{blue}{crossover} and \textcolor{red}{mutation} to produce a new population. Complete list of mutation operators is provided in Table \ref{mutation_op_tbl}}
   \label{code_mutation_algs}
\end{figure}

\begin{table*}[t]
  \centering
  \footnotesize
  \begin{tabular}{p{0.15\linewidth} p{0.15\linewidth} p{0.03\linewidth} p{0.55\linewidth}}
    Operator & Allowed Functions & Prob & Description\\ \cline{1-4}
    Insert Instruction & \textcode{GetAction()} \textcode{CADF()} & 0.5 & Insert randomly generated instruction at uniformly sampled line number \\
    Delete Instruction & \textcode{GetAction()} \textcode{CADF()} & 1.0 & Delete the instruction at a uniformly sampled line number \\
    Randomize Instruction & \textcode{GetAction()} \textcode{CADF()} & 1.0 & Randomize the instruction at a uniformly sampled line number \\
    Randomize Function & \textcode{GetAction()} \textcode{CADF()} & 0.1 & Randomly shuffles all lines of code \\
    Randomize constants & \textcode{StartEpisode()} & 0.5 & Modify a fraction (0.2) of uniformly sampled constants in a uniformly sampled instruction. For each constant, add noise sampled from $\mathcal{N}(0, 0.05^{2})$. \\
    Randomize Parameter & \textcode{GetAction()} \textcode{CADF()} & 0.5 & Randomize a uniformly sampled parameter in a uniformly sampled instruction \\
    Randomize dim indices & \textcode{GetAction()} \textcode{CADF()} & 0.5 & Randomize a fraction (0.2) of uniformly sampled dim indices in a uniformly sampled instruction. Each chosen dim index is set to a new integer uniformly sampled from $[0, dim)$ where $dim$ is the size of the memory structure being referenced.
  \end{tabular}
  \caption{Mutation operators. \textit{Prob} column lists the relative probability of applying each operation. For example, the Delete Instruction op will be applied twice as often as the Insert instruction.}
  \label{mutation_op_tbl}
\end{table*}

\subsection{Baseline Details}\label{baseline_supp}
\textit{Augmented Random Search (ARS):} We used a standard implementation from \cite{NEURIPS2018_7634ea65} and hyperparameter tuned over a cross product between:
\begin{itemize}
\item learning rate: [0.001, 0.005, 0.01, 0.05, 0.1, 0.5]
\item Gaussian standard deviation: [0.001, 0.005, 0.01, 0.05, 0.1, 0.5]
\end{itemize}
and used a 2-layer MLP of hidden layer sizes (32,32) with Tanh non-linearity, along with an LSTM of size 32.

\textit{Proximal Policy Optimization (PPO):} We used a standard implementation from TF-Agents \cite{tf_agents}, which we verified to reproduce standard Mujoco results from \cite{ppo}. We varied the following hyperparameters:
\begin{itemize}
\item nsteps ("collect sequence length"): [256, 1024]
\item learning rate: [5e-5, 1e-4, 5e-4, 1e-3, 5e-3]
\item entropy regularization: [0.0, 0.05, 0.1, 0.5]
\end{itemize}
and due to the use of a shared value network, we used a 2-layer MLP of hidden layer sizes (256, 256) with ReLU nonlinearity alongside an LSTM of size 256. Since PPO significantly underperformed (e.g., obtaining only $\approx$100 reward on quadruped tasks), we omitted its results in this paper to save space.

\subsection{Quadruped Tasks}\label{quadruped_task_supp}
We perform 20 independent repeats for each method with unique random seeds. All repeats are allowed to train until convergence. NSGA-II uses parent and child population sizes of 100 and 1000, respectively. No search restarts or FEC are enabled. The set of operations available for inclusion in any program are listed in Table \ref{ops_table}. For ARS experiments, we run a hyperparameter sweep consisting of 36 repeats with unique hyperparameters. We then run an additional 20 repeats using the best hyperparameter configuration.

\begin{figure}[!h]
	\centering
	\includegraphics[width=0.25\textwidth]{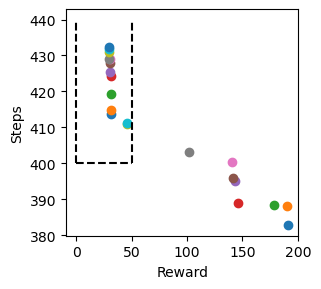}
	\caption{A typical Pareto front early in NSGA-II search. The dashed box shows policies that are effectively eliminated through fitness constraints.}
	\label{nsga2_constraints}
\end{figure}

\subsection{Cataclysmic Cartpole Tasks}\label{cartpole_task_supp}

Cartpole \cite{sutton18, brockman2016gym} is a classic control task in which a pole is attached by an un-actuated joint to a cart that moves Left or Right along a frictionless track, Figure \ref{cartpole_fig}. The observable state of the system at each timestep, $\vec{s}(t)$, is described by 4 variables including the cart position ($x$), cart velocity ($\dot{x}$), pole angle relative to the cart ($\theta$), and pole angular velocity ($\dot{\theta}$). We use a continuous-action version of the problem in which the system is controlled by applying a force $\in [-1, 1]$ to the cart. The pole starts nearly upright, and the goal is to prevent it from falling over. An episode ends when the pole is more than 12 degrees from vertical, the cart moves more than 2.4 units from the center, or a time constraint is reached (1000 timesteps). A reward of $(1 - |\theta_{vert}|/12)^2$ is provided for every timestep that the pole remains upright, where $\theta_{vert}$ is a fixed reference for the angle of the pole relative to the vertical plane. As such, the objective is to balance the pole close to vertical for as long as possible. 

\begin{figure}[!h]
	\centering
	\includegraphics[width=0.47\textwidth]{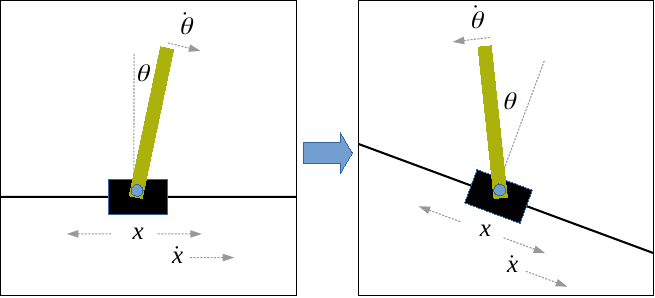}
	\caption{Illustration of a track angle change in the Cataclysmic Cartpole task with the 4 variables in the state observation $\vec{s}(t)$. Note that $\theta$ always represents the angle between the pole and the line running perpendicular to the track and cart, thus the \textit{desired} value of $\theta$ to maintain balance $(\theta_{vert} = 0)$ changes with the track angle and is not directly observable to the policy.}
	\label{cartpole_fig}
\end{figure}

\begin{enumerate}
    \item Sudden: A sudden change in each change parameter occurs at a unique random timestep in [200, 800], Figure \ref{change_sudden_fig}.
    \item Continuous: Each parameter changes over a window with random, independently chosen start and stop timesteps in [200, 800], Figure \ref{change_continuous_fig}.
\end{enumerate}

\begin{figure}[!h]
     \centering
     \begin{subfigure}[b]{0.48\textwidth}
         \centering
         \includegraphics[width=\textwidth]{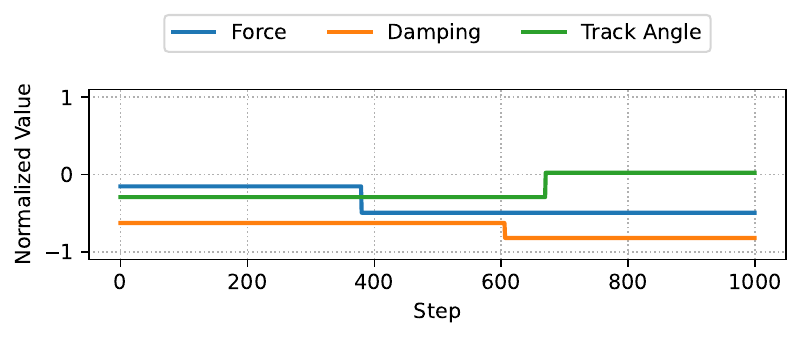}
         \caption{Sudden}
         \label{change_sudden_fig}
     \end{subfigure}
     \begin{subfigure}[b]{0.48\textwidth}
         \centering
         \includegraphics[width=\textwidth]{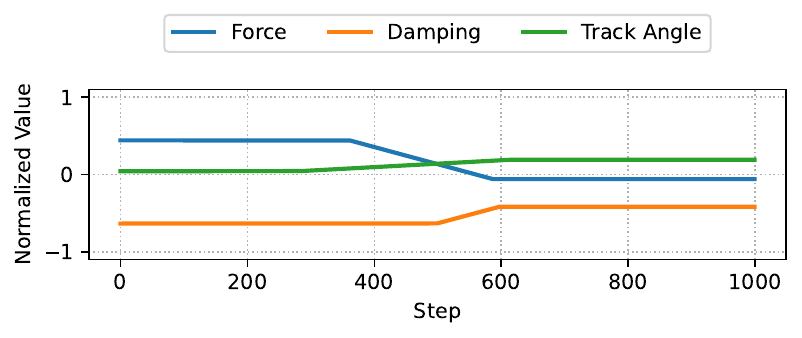}
         \caption{Continuous}
         \label{change_continuous_fig}
     \end{subfigure}
        \caption{A typical randomly-created change schedule.}
        \label{change_schedule_fig}
\end{figure}

For the ARZ methods, we execute 10 repeats of each experiment with unique random seeds. For ARS, we run a hyperparameter sweep consisting of 36 repeats with unique hyperparameters. In each case, we select 5 repeats with the best search fitness and test the single best policy from each. 
Plots show mean fitness over 100 episodes for each policy in each task.

} 

\supplsection{Additional Experiments: Cataclysmic Cartpole}{ap:cartpole_experiments}{
\subsection{Adaptation Gap}\label{adaptation_gap_results_sec}
In this section we use stateless policies (ARZ and MLP) to confirm that Cataclysmic Cartpole dynamics are significant enough to create an \textbf{adaptation gap}: when stateless policies (i.e. generalists) fail to perform well because they cannot adapt their control policy in the non-stationary environment. As mentioned in Section \ref{temporal_memory_sec} our evolved algorithms are able to adapt partly because they are stateful: the contents of their memory (\textcode{sX}, \textcode{vX}, \textcode{mX}, and \textcode{iX}) are persistent across timesteps of an episode. The representation can easily support stateless algorithms simply by forcing the policy to wipe its memory content and re-initialize constants at the beginning of the \textcode{GetAction()} function (See Figure \ref{algorithm_evaluation_fig}). 

 Fig. \ref{cartpole_metaval_continuous_baselines_fig} indicates that, in the continuous change environment, the stateless baselines (MLP and ARZ stateless) fail to achieve sufficient fitness ($\approx800$) when all types of change occur simultaneously (\textcode{ALL}). This confirms that the continuous change paradigm provides a suitably challenging non-stationary problem environments to study adaptation and life-long learning. In the sudden change task (Figure \ref{cartpole_metaval_sudden_baselines_fig}), the MLP baseline still fails. Surprizingly, ARZ can discover stateless policies that succeed under this type of non-stationarity. 
 
\begin{figure}[!h]
   \centering
   \includegraphics[width=0.48\textwidth]{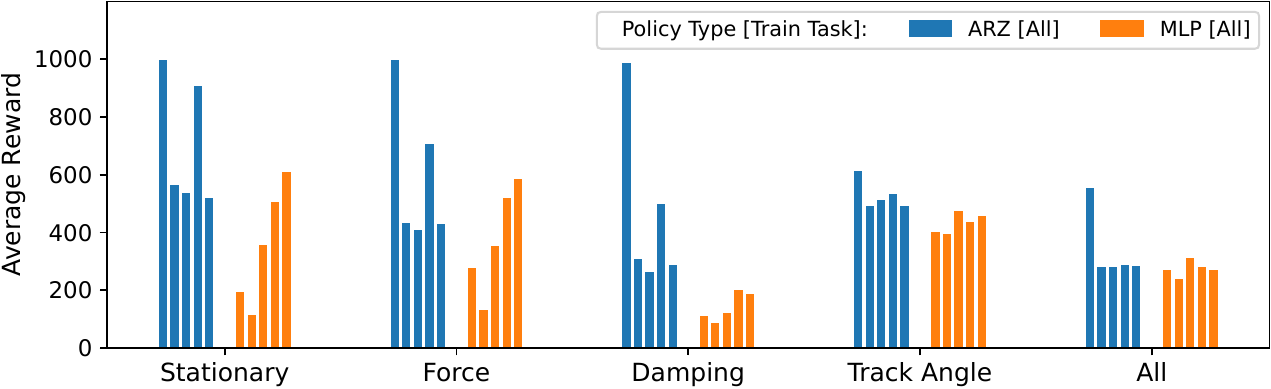}
   \caption{Stateless baselines fail to achieve sufficient fitness ($\approx800$) when all types of change occur simultaneously (\textcode{ALL}). The plot shows test results for \textbf{stateless baselines} in the Cataclysmic Cartpole \textbf{continuous change} tasks. The legend indicates policy type and search task. "Stationary" is the standard Cartpole task while "Force", "Damping", and "Track Angle" refer to Cartpole with continuous change in these parameters \textit{only} (See Section \ref{cartpole_sec}). "All" is the case where all change parameters are potentially changing simultaneously. Y-axis is the average reward of 100 episodes in each task. See Section \ref{adaptation_gap_results_sec} for discussion.}
   \label{cartpole_metaval_continuous_baselines_fig}
\end{figure}

\begin{figure}[!h]
   \centering
   \includegraphics[width=0.48\textwidth]{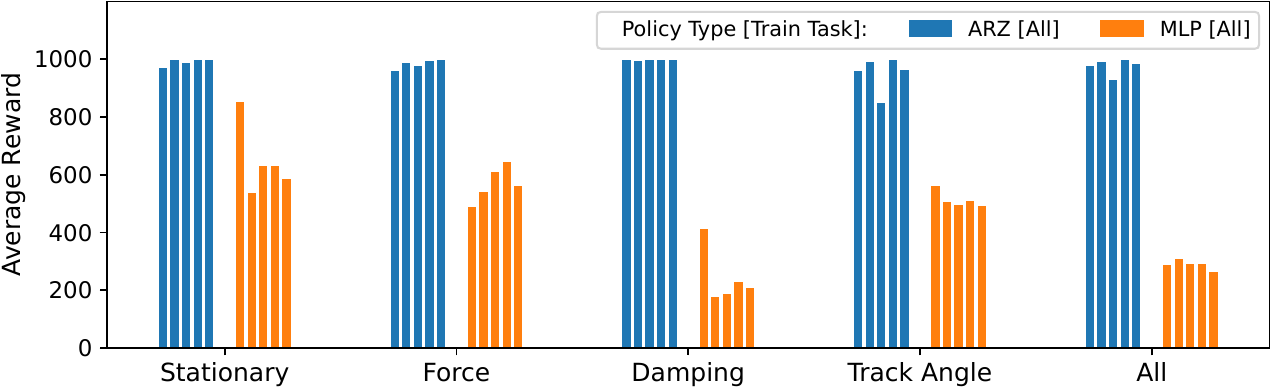}
   \caption{ARZ can discover stateless policies that succeed in the sudden change tasks. The plot shows test results for \textbf{stateless baselines} in the Cartpole \textbf{sudden change} tasks. The legend indicates policy type and search task. "Stationary" is the standard Cartpole task while "Force", "Damping", and "Track Angle" refer to Cartpole with continuous change in these parameters \textit{only} (See Section \ref{cartpole_sec}). "All" is the case where all change parameters are potentially changing simultaneously. Y-axis is the average reward of 100 episodes in each task. See Section \ref{adaptation_gap_results_sec} for discussion.}
   \label{cartpole_metaval_sudden_baselines_fig}
\end{figure}

\subsection{Adapting to Unseen Dynamics in Cataclysmic Cartpole}\label{unanticipated_changes_results_sec}
How can we build adaptive control policies without any prior knowledge about what type of environmental change might occur? Surprisingly, for ARZ, we find that injecting partial-observability and dynamic actuator noise during evolution (training) can act as a general surrogate for non-stationary task dynamics, supporting the emergence of policies that can adapt to novel task dynamics that were \textit{not} experienced during evolution. This was not possible for our LSTM baselines. It is a significant finding that deserves more attention in future work because it implies we can potentially evolve proficient control policies without complete prior knowledge of their task environment dynamics, thus relaxing the need for an accurate physics simulator.

If we assume that no simulator is available for any of the non-stationary tasks in Cataclysmic Cartpole (Force, Damping, Track Angle), can we still build policies that cope with these changes? From a policy's perspective, changes to the physics of the environment will (1) change the meaning of its sensor observations (e.g. pole angle sensor value ($\theta$) corresponding to vertical suddenly changes); and/or (2) change the effect of its actions (e.g. a particular actuator value suddenly has a much greater effect on the cart's trajectory). To prepare policies for these uncertainties, we evolve them with non-stationary noise applied to their actions and introduce a partially-observable observation space. Specifically, we modify the task to add:
\begin{itemize}
\item Actuator Noise: Each action value $v$ is modified such that $v = v + n$, where $n$ is sampled from a Gaussian distribution with mean that varies in [-2, 2] following the continuous change schedule in Figure \ref{change_continuous_fig}. 
\item Partial Observability: Positional state variables (cart position ($x$) and pole angle relative to the cart ($\theta$)) are set to zero prior to passing the state observation to the policy. 
\end{itemize}

Our hypothesis is that this will encourage policies to rely less on their observations and actions, and as a result they might build a stronger, more dynamic internal world-model to predict how their actions will affect future states. That is, there is more pressure to model the environment's dynamic transition function. In Figure \ref{cartpole_metaval_continous_b_fig}, \textcode{ARZ [PO + Act Noise]} shows test results for an ARZ experiment that uses the stationary task simulator during evolution (i.e. the unmodified Cartpole environment) but applies actuator noise and partial observability as described above. Remarkably, these evolved policies are able to adapt reasonably well under all non-stationary tasks in the Cataclysmic Cartpole environment, achieving an average reward of $\geq 700$ in all tasks. Using the same search configuration, ARS does not discover parameters for an LSTM network that supports adaptation to all non-stationary tasks (\textcode{LSTM [PO + Act Noise]}).  

\begin{figure}[!h]
	\centering
	\includegraphics[width=0.48\textwidth]{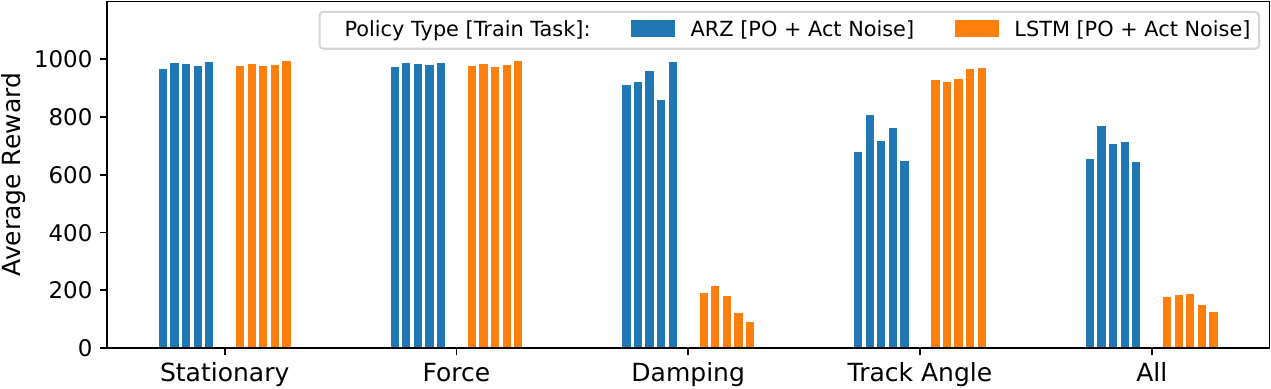}
	\caption{ARZ can discover policies that adapt to unseen tasks. The plot shows post-evolution test results for \emph{adapting policies} in the Cartpole \emph{continuous change} tasks. The legend indicates policy type and search task. \textcode{[All]} indicate policies were exposed to all tasks during evolution. \textcode{[PO + Act Noise]} indicates policies were evolved with partial observability and action noise on the stationary task, while the dynamic change tasks were unseen until test. Y-axis is the average reward of 100 episodes in each task. See Section \ref{unanticipated_changes_results_sec} for discussion.}
	\label{cartpole_metaval_continous_b_fig}
\end{figure}

In summary, preliminary data presented in this section suggests that adding partial-observability and actuator noise to the stationary Cartpole task during search allows ARZ to discover policies that can adapt to \textit{unseen} non-stationary tasks, a methodology that does not work for ARS with LSTM networks. We leave comprehensive analysis of these findings to future work.

} 

\supplsection{Cartpole Algorithm Analysis}{ap:algo_analysis}{

Here we analyze the algorithm presented in Figure \ref{sample_stateful_all_algo}:
\begin{figure}[!h]
    \begin{code}{0.48\textwidth}{1.3in}{codebackground}
        \codeline{\codecomment{sX: scalar memory at address X.}}
        \codeline{\codecomment{obs: vector [x, theta, x\_dot, theta\_dot].}}
        \codeline{\codecomment{a, b, c: fixed scalar parameters.}}
        \codeline{\codecomment{V, W: 4-dimensional vector parameters.}}
        \codeline{\codedef{def} GetAction(obs, action):}
        \codeline{\codetab s0 = a * s2 + action}
        \codeline{\codetab s1 = s0 + s1 + b * action + dot(V, obs)}
        \codeline{\codetab s2 = s0 + c * s1}
        \codeline{\codetab action = s0 + dot(obs, W)}
        \codeline{\codetab \codereturn{return} action}
    \end{code}
    \caption{Sample stateful action function evolved on the Cataclysmic Cartpole task where all parameters are subject to continuous change (ARZ [All] in Fig.\ \ref{cartpole_metaval_continous_a_fig}). Code shown in Python. This figure is a repeat of Figure \ref{sample_stateful_all_algo}.}
\end{figure}

Organizing the values of $\mu=\text{\textcode{s0}}$, $\nu=\text{\textcode{s1}}$ and $\xi=\text{\textcode{s2}}$ at step $n$ into a vector:
\begin{equation}
Z_n = (\mu_{n},~\nu_{n+1},~\xi_{n})^T \,, \nonumber
\end{equation}
and concatenating the observation vector at step $n+1$ and the action at step $n$ into a state vector $s_{n+1}$:
\begin{equation}
s_{n+1} = (x_{n+1},~\theta_{n+1},~\dot{x}_{n+1},~\dot{\theta}_{n+1},~ \text{act}_{n})^T \,, \nonumber
\end{equation}
we can re-write the value of these accumulators at step $n$ in the following way:
\begin{align}
s_{n+1} &= \text{concat}(\text{obs}_{n+1}\,,~\text{act}_n) \nonumber \\
Z_{n+1} &= \widetilde{U} \cdot Z_n + \widetilde{P} \cdot s_{n+1} \,, \label{recurrent}\\
\text{act}_{n+1} &= \widetilde{A}^T \cdot Z_{n+1} + \widetilde{W}^T \cdot s_{n+1} \,.
\nonumber
\end{align}
The particular variables used in this formula map to the parameters of Figure \ref{sample_stateful_all_algo} as follows:
\begin{align}
\widetilde{U} &= \begin{pmatrix}
0 & 0 & a \\ 0 & 1 & a \\ 0 & c & a(1+c)
\end{pmatrix} \,, \nonumber\\
\widetilde{P} &=
\begin{pmatrix}
0 & 0 & 0 & 0 & 1 \\
V_1 & V_2 & V_3 & V_4 & b+1 \\
cV_1 & cV_2 & cV_3 & cV_4 & a + bc + c
\end{pmatrix} \,, \nonumber\\
\widetilde{A} &=
(1,~0,~0)^T \,, \nonumber\\
\widetilde{W} &=
(W_1,~W_2,~W_3,~W_4,~0)^T \,. \nonumber
\end{align}
The numerical values of the parameters of the model found are given by
\begin{align}
a &= -0.549 \,,\quad b= -0.673 \,, \quad c= 0.082 \,, \nonumber \\
V &=  (-1.960,~-0.7422,~0.7373,~-5.284)^T \,, \nonumber \\
W &=  (0.0,~0.365,~2.878,~2.799)^T \,. \nonumber
\end{align}

Equation (\ref{recurrent}) can be viewed as a linear recurrent model, where $Z_n$ is the internal state of the model. The action at the $n$-th step is obtained as a linear function of the internal state, the observation vector and the action value at the previous step. An interesting aspect of the particular model found is that the matrix $\widetilde{U}$ by construction has eigenvalues $0$, $1$ and $a(1+c) \approx -0.594$.

Equation (\ref{recurrent}) being a simple linear model, we may write $\text{act}_{n+1}$ explicitly as a sum:
\begin{align}
\text{act}_n =& \widetilde{A}^T \cdot \widetilde{U}^{n} \cdot Z_0 + \widetilde{W}^T \cdot s_{n}  \nonumber \\
&+ \widetilde{A}^T \cdot
\sum_{i = 0}^{n} \widetilde{U}^{n-i} \cdot \widetilde{P} \cdot s_i \,.  \nonumber
\end{align}
When taking the continuous limit of this expression, there is a subtlety in that the $s_{n+1}$ vector is obtained by composing the observation vector at time step $n+1$ and the action value at time step $n$. We can nevertheless be careful to redefine $s$ to be made up of concurrent components and still arrive at an expression which in the continuous limit, takes the form:
\begin{align}
\text{act}(t) =& c + w^T \cdot U^t \cdot u + p^T \cdot s(t) \nonumber \\
&+ v^T \cdot \int_0^t du ~ U^{t-u} \cdot P \cdot s(u) \label{genpid} \,.
\end{align}
We note that when we set $U = \text{Id}$ this expression straightforwardly reduces to a PID controller-like model:
\begin{align}
\text{act}(t) =& (c + w^T \cdot u) + p^T \cdot s(t) + (v^T \cdot P) \cdot \int_0^t du s(u) \nonumber \,.
\end{align}

An instructive way of re-writing Equation (\ref{genpid}) is to explicitly use the eigenvalues $e^{-\omega_k}$ of $U$. The equation can be re-parameterized as
\begin{align}
\text{act}(t) =& c + \sum_{k=1}^{d} c_k e^{-\omega_k t}  + p^T \cdot s(t) \nonumber \\
&+ \sum_{k=1}^d v_k^T \cdot \int_0^t du ~ e^{-\omega_k (t-u)} s(u) \nonumber\,.
\end{align}
Here it is clear that the expression is a straightforward generalization of the PID controller, where only the weight-one cumulant of the history is utilized to compute the action. Now, a multitude of cumulants with distinct decay rates can be utilized.

} 

\supplsection{Complexity Comparison}{ap:complexity}{

\subsection{Baselines}

As noted in section \ref{ap:methods}, MLP and LSTM networks have been trained with ARS as baselines for the adaptation tasks in the paper. We can estimate a lower bound for the number parameters and floating point operations required for each model by only counting the matrix variables for the parameter count and matrix multiplications for the floating point operations. This negelects the bias variables and non-matrix multiplication ops such as application of non-linearities or vector component-wise multiplications.

Given the input dimension $d_\text{in}$, the output dimension $d_\text{out}$ and the internal dimension $d$, we find that the number of parameters and the floating point operations for the MLP and LSTM model per step is given by:
\begin{align}
\text{FLOPS}_\text{MLP} &\approx 2 \times \text{Params}_\text{MLP} > 2d(d_\text{in} + d + d_\text{out}) \\
\text{FLOPS}_\text{LSTM} &\approx 2 \times \text{Params}_\text{LSTM} > 2d(4d_\text{in} + 4d + d_\text{out})
\end{align}

\subsection{Quadruped Leg-Breaking}
The algorithm presented in Figure \ref{quadruped_algo} contains $16 + 16 \times 37 = 608$ parameters and executes a maximum of $54 \times 37 + 82 = 2080$ floating point ops per step, where we have counted all operations acting on floats or pairs of floats, assuming that all of the ``if" statements pass. The input and output dimensions of the tasks are 37 and 12, while the ARS-trained models have internal dimensions $d=32$. Using the formulae above, we see that the MLP model contains over 2592 parameters and uses more than 5184 FLOPs. Meanwhile the LSTM model uses more than 9216 parameters and 18432 FLOPs.

\subsection{Cataclysmic Cartpole}
The algorithm presented in Figure \ref{cartpole_metaval_continous_a_fig} contains 11 parameters and executes 25 floating point ops per step. The input and output dimensions of the tasks are 4 and 1, with internal dimensions $d=32$ for the neural networks. The MLP model contains over 1184 parameters and uses more than 2368 FLOPs. The LSTM model uses more than 4604 parameters and 9280 FLOPs.

\subsection{Discussion}
The efficiency of ARZ policies stems from two characteristics of the system. First, like many genetic programming methods, ARZ builds policies starting from simple algorithms and incrementally adds complexity through interaction with the task environment (e.g., \cite{koza1990,brameier07}). This implies that the computational cost of action inference is low early in evolution, and only increases as more complex structures provide fitness gains. In other words, the search is bound by \emph{incremental growth}. Second, in ARZ, mutation is twice as likely to remove an instruction than insert an instruction (See Table \ref{mutation_op_tbl}), which has been found to have a regularization effect on the population \cite{real2020automl}.
}  

\supplsection{Search Space Additional Details}{ap:search_space_additional_details}{

Supplementary Table~\ref{ops_table} describes the set of operations in our search space. Note that no matrix operations were used for the quadruped robot domain.

} 

\newcommand{\sepforall}{\:\forall}
\newcommand{\rowsep}{2pt}

\begin{table*}[!hb]
\caption{Ops vocabulary. $s$, $\vec{v}$ and $M$ denote a scalar, vector, and matrix, resp. Early-alphabet letters ($a$, $b$, \etc) denote memory addresses. Mid-alphabet letters (\eg $i$, $j$, \etc) denote vector/matrix indexes (``Index'' column). Greek letters denote constants (``Consts.'' column). $\mathcal{U}(\alpha,\beta)$ denotes a sample from a uniform distribution in $[\alpha,\beta]$. $\mathcal{N}(\mu,\sigma)$ is analogous for a normal distribution with mean $\mu$ and standard deviation $\sigma$. $\mathbbm{1}_X$ is the indicator function for set $X$. Example: ``$M_a^{(i,j)} = \mathcal{U}(\alpha,\beta)$'' describes the operation ``assign to the $i$,$j$-th entry of the matrix at address $a$ a value sampled from a uniform random distribution in $[\alpha,\beta]$''.}
\label{ops_table}
\begin{center}
\begin{scriptsize}
\begin{tabular}{p{0.3in}|
                >{\centering\arraybackslash}p{1.3in}|
                >{\centering\arraybackslash}p{0.7in} >{\centering\arraybackslash}p{0.33in}|
                >{\centering\arraybackslash}p{0.5in} >{\centering\arraybackslash}p{0.26in}|
                >{\centering\arraybackslash}p{1.98in}l}
\cmidrule(r{8pt}){1-7}
Op & Code    & \multicolumn{2}{c|}{Input Args} & \multicolumn{2}{c|}{Output Args} & Description   & \\
ID & Example & Addresses & Consts.       & Address    & Index                     & (see caption) & \\
  &          & / types   &               & / type     &                           &               & \\
\cmidrule(r{8pt}){1-7}
OP1 & \textcode{no\_op} & -- & -- & -- & -- & -- & \\[\rowsep]
OP2 & \textcode{s2=s3+s0} & $a$,$b$ / scalars & -- & $c$ / scalar & -- & $s_c = s_a + s_b$ & \\[\rowsep]
OP3 & \textcode{s4=s0-s1} & $a$,$b$ / scalars & -- & $c$ / scalar & -- & $s_c = s_a - s_b$ & \\[\rowsep]
OP4 & \textcode{s8=s5*s5} & $a$,$b$ / scalars & -- & $c$ / scalar & -- & $s_c = s_a \, s_b$ & \\[\rowsep]
OP5 & \textcode{s7=s5/s2} & $a$,$b$ / scalars & -- & $c$ / scalar & -- & $s_c = s_a / s_b$ & \\[\rowsep]
OP6 & \textcode{s8=abs(s0)} & $a$ / scalar & -- & $b$ / scalar & -- & $s_b = |s_a|$ & \\[\rowsep]
OP7 & \textcode{s4=1/s8} & $a$ / scalar & -- & $b$ / scalar & -- & $s_b = 1/s_a$ & \\[\rowsep]
OP8 & \textcode{s5=sin(s4)} & $a$ / scalar & -- & $b$ / scalar & -- & $s_b = \sin(s_a)$ & \\[\rowsep]
OP9& \textcode{s1=cos(s4)} & $a$ / scalar & -- & $b$ / scalar & -- & $s_b = \cos(s_a)$ & \\[\rowsep]
OP10 & \textcode{s3=tan(s3)} & $a$ / scalar & -- & $b$ / scalar & -- & $s_b = \tan(s_a)$ & \\[\rowsep]
OP11 & \textcode{s0=arcsin(s4)} & $a$ / scalar & -- & $b$ / scalar & -- & $s_b = \arcsin(s_a)$ & \\[\rowsep]
OP12 & \textcode{s2=arccos(s0)} & $a$ / scalar & -- & $b$ / scalar & -- & $s_b = \arccos(s_a)$ & \\[\rowsep]
OP13 & \textcode{s4=arctan(s0)} & $a$ / scalar & -- & $b$ / scalar & -- & $s_b = \arctan(s_a)$ & \\[\rowsep]
OP14 & \textcode{s1=exp(s2)} & $a$ / scalar & -- & $b$ / scalar & -- & $s_b = e^{s_a}$ & \\[\rowsep]
OP15 & \textcode{s0=log(s3)} & $a$ / scalar & -- & $b$ / scalar & -- & $s_b = \operatorname{log}{s_a}$ & \\[\rowsep]
OP16 & \textcode{s3=heaviside(s0)} & $a$ / scalar & -- & $b$ / scalar & -- & $s_b = \mathbbm{1}_{\mathbb{R}^+}(s_a)$ & \\[\rowsep]
OP17 & \textcode{v2=heaviside(v2)} & $a$ / vector & -- & $b$ / vector & -- & $\vec{v}_b^{\,(i)} = \mathbbm{1}_{\mathbb{R}^+}(\vec{v}_a^{\,(i)}) \sepforall i$ & \\[\rowsep]
OP18 & \textcode{m7=heaviside(m3)} & $a$ / matrix & -- & $b$ / matrix & -- & $M_b^{(i,j)} = \mathbbm{1}_{\mathbb{R}^+}(M_a^{(i,j)}) \sepforall i,j$ & \\[\rowsep]
OP19 & \textcode{v1=s7*v1} & $a$,$b$ / sc,vec & -- & $c$ / vector & -- & $\vec{v}_c = s_a \, \vec{v}_b$ & \\[\rowsep]
OP20 & \textcode{v1=bcast(s3)} & $a$ / scalar & -- & $b$ / vector & -- & $\vec{v}_b^{\,(i)} = s_a \ \ \sepforall i$ & \\[\rowsep]
OP21 & \textcode{v5=1/v7} & $a$ / vector & -- & $b$ / vector & -- & $\vec{v}_b^{\,(i)} = 1/\vec{v}_a^{\,(i)} \ \ \sepforall i$ & \\[\rowsep]
OP22 & \textcode{s0=norm(v3)} & $a$ / scalar & -- & $b$ / vector & -- & $s_b = |\vec{v}_a|$ & \\[\rowsep]
OP23 & \textcode{v3=abs(v3)} & $a$ / vector & -- & $b$ / vector & -- & $\vec{v}_b^{\,(i)} = |\vec{v}_a^{\,(i)}| \ \ \sepforall i$ & \\[\rowsep]
OP24 & \textcode{v5=v0+v9} & $a$,$b$ / vectors & -- & $c$ / vector & -- & $\vec{v}_c = \vec{v}_a + \vec{v}_b$ & \\[\rowsep]
OP25 & \textcode{v1=v0-v9} & $a$,$b$ / vectors & -- & $c$ / vector & -- & $\vec{v}_c = \vec{v}_a - \vec{v}_b$ & \\[\rowsep]
OP26 & \textcode{v8=v1*v9} & $a$,$b$ / vectors & -- & $c$ / vector & -- & $\vec{v}_c^{\,(i)} = \vec{v}_a^{\,(i)} \, \vec{v}_b^{\,(i)} \sepforall i$ & \\[\rowsep]
OP27 & \textcode{v9=v8/v2} & $a$,$b$ / vectors & -- & $c$ / vector & -- & $\vec{v}_c^{\,(i)} = \vec{v}_a^{\,(i)} / \vec{v}_b^{\,(i)} \sepforall i$ & \\[\rowsep]
OP28 & \textcode{s6=dot(v1,v5)} & $a$,$b$ / vectors & -- & $c$ / scalar & -- & $s_c=\vec{v}_a^{\,T} \, \vec{v}_b$ & \\[\rowsep]
OP29 & \textcode{m1=outer(v6,v5)} & $a$,$b$ / vectors & -- & $c$ / matrix & -- & $M_c=\vec{v}_a \, \vec{v}_b^{\,T}$ & \\[\rowsep]
OP30 & \textcode{m1=s4*m2} & $a$,$b$ / sc/mat & -- & $c$ / matrix & -- & $M_c=s_a \, M_b$ & \\[\rowsep]
OP31 & \textcode{m3=1/m0} & $a$ / matrix & -- & $b$ / matrix & -- & $M_b^{(i,j)}=1/M_a^{(i,j)} \sepforall i,j$ & \\[\rowsep]
OP32 & \textcode{v6=dot(m1,v0)} & $a$,$b$ / mat/vec & -- & $c$ / vector & -- & $\vec{v}_c = M_a \, \vec{v}_b$ & \\[\rowsep]
OP33 & \textcode{m2=bcast(v0,axis=0)} & $a$ / vector & -- & $b$ / matrix & -- & $M_b^{(i,j)} = \vec{v}_a^{\,(i)} \sepforall i,j$ & \\[\rowsep]
OP34 & \textcode{m2=bcast(v0,axis=1)} & $a$ / vector & -- & $b$ / matrix & -- & $M_b^{(j,i)} = \vec{v}_a^{\,(i)} \sepforall i,j$ & \\[\rowsep]
OP35 & \textcode{s2=norm(m1)} & $a$ / matrix & -- & $b$ / scalar & -- & $s_b = ||M_a||$ & \\[\rowsep]
OP36 & \textcode{v4=norm(m7,axis=0)} & $a$ / matrix & -- & $b$ / vector & -- & $\vec{v}_b^{\,(i)} = |M_a^{(i,\cdot)}| \sepforall i$ & \\[\rowsep]
OP37 & \textcode{v4=norm(m7,axis=1)} & $a$ / matrix & -- & $b$ / vector & -- & $\vec{v}_b^{\,(j)} = |M_a^{(\cdot,j)}| \sepforall j$ & \\[\rowsep]
\multicolumn{7}{c}{\dotfill [Table continues on the next page.] \dotfill}\\
\end{tabular}
\end{scriptsize}
\end{center}
\end{table*}

\addtocounter{table}{-1}
\begin{table*}[!t]
\vspace*{20pt}
\caption{Ops vocabulary (continued)}
\begin{center}
\begin{scriptsize}
\begin{tabular}{p{0.3in}|
                >{\centering\arraybackslash}p{1.3in}|
                >{\centering\arraybackslash}p{0.7in} >{\centering\arraybackslash}p{0.33in}|
                >{\centering\arraybackslash}p{0.5in} >{\centering\arraybackslash}p{0.26in}|
                >{\centering\arraybackslash}p{1.98in}l}
\cmidrule(r{8pt}){1-7}
Op & Code    & \multicolumn{2}{c|}{Input Args} & \multicolumn{2}{c|}{Output Args} & Description   & \\
ID & Example & Addresses & Consts       & Address    & Index                     & (see caption) & \\
  &          & / types   &              & / type     &                           &               & \\
\cmidrule(r{8pt}){1-7}
OP38 & \textcode{m9=transpose(m3)} & $a$ / matrix & -- & $b$ / matrix & -- & $M_b = |M_a^T|$ & \\[\rowsep]
OP39 & \textcode{m1=abs(m8)} & $a$ / matrix & -- & $b$ / matrix & -- & $M_b^{(i,j)} = |M_a^{(i,j)}| \sepforall i,j$ & \\[\rowsep]
OP40 & \textcode{m2=m2+m0} & $a$,$b$ / matrixes & -- & $c$ / matrix & -- & $M_c = M_a + M_b$ & \\[\rowsep]
OP41 & \textcode{m2=m3-m1} & $a$,$b$ / matrixes & -- & $c$ / matrix & -- & $M_c = M_a - M_b$ & \\[\rowsep]
OP42 & \textcode{m3=m2*m3} & $a$,$b$ / matrixes & -- & $c$ / matrix & -- & $M_c^{(i,j)} = M_a^{(i,j)} \, M_b^{(i,j)} \sepforall i,j$ & \\[\rowsep]
OP43 & \textcode{m4=m2/m4} & $a$,$b$ / matrixes & -- & $c$ / matrix & -- & $M_c^{(i,j)} = M_a^{(i,j)} / M_b^{(i,j)} \sepforall i,j$ & \\[\rowsep]
OP44 & \textcode{m5=matmul(m5,m7)} & $a$,$b$ / matrixes & -- & $c$ / matrix & -- & $M_c = M_a \, M_b$ & \\[\rowsep]
OP45 & \textcode{s1=minimum(s2,s3)} & $a$,$b$ / scalars & -- & $c$ / scalar & -- & $s_c = \min(s_a, s_b)$ & \\[\rowsep]
OP46 & \textcode{v4=minimum(v3,v9)} & $a$,$b$ / vectors & -- & $c$ / vector & -- & $\vec{v}_c^{\,(i)} = \min(\vec{v}_a^{\,(i)}, \vec{v}_b^{\,(i)}) \sepforall i$ & \\[\rowsep]
OP47 & \textcode{m2=minimum(m2,m1)} & $a$,$b$ / matrixes & -- & $c$ / matrix & -- & $M_c^{(i,j)} = \min(M_a^{(i,j)}, M_b^{(i,j)}) \sepforall i,j$ & \\[\rowsep]
OP48 & \textcode{s8=maximum(s3,s0)} & $a$,$b$ / scalars & -- & $c$ / scalar & -- & $s_c = \max(s_a, s_b)$ & \\[\rowsep]
OP49 & \textcode{v7=maximum(v3,v6)} & $a$,$b$ / vectors & -- & $c$ / vector & -- & $\vec{v}_c^{\,(i)} = \max(\vec{v}_a^{\,(i)}, \vec{v}_b^{\,(i)}) \sepforall i$ & \\[\rowsep]
OP50 & \textcode{m7=maximum(m1,m0)} & $a$,$b$ / matrixes & -- & $c$ / matrix & -- & $M_c^{(i,j)} = \max(M_a^{(i,j)}, M_b^{(i,j)}) \sepforall i,j$ & \\[\rowsep]
OP51 & \textcode{s2=mean(v2)} & $a$ / vector & -- & $b$ / scalar & -- & $s_b = \operatorname{mean}(\vec{v}_a)$ & \\[\rowsep]
OP52 & \textcode{s2=mean(m8)} & $a$ / matrix & -- & $b$ / scalar & -- & $s_b = \operatorname{mean}(M_a)$ & \\[\rowsep]
OP53 & \textcode{v1=mean(m2,axis=0)} & $a$ / matrix & -- & $b$ / vector & -- & $\vec{v}_b^{\,(i)} = \operatorname{mean}(M_a^{(i,\cdot)}) \sepforall i$ & \\[\rowsep]
OP54 & \textcode{v3=std(m2,axis=0)} & $a$ / matrix & -- & $b$ / vector & -- & $\vec{v}_b^{\,(i)} = \operatorname{stdev}(M_a^{(i,\cdot)}) \sepforall i$ & \\[\rowsep]
OP55 & \textcode{s3=std(v3)} & $a$ / vector & -- & $b$ / scalar & -- & $s_b = \operatorname{stdev}(\vec{v}_a)$ & \\[\rowsep]
OP56 & \textcode{s4=std(m0)} & $a$ / matrix & -- & $b$ / scalar & -- & $s_b = \operatorname{stdev}(M_a)$ & \\[\rowsep]
OP57 & \textcode{s2=C1} & -- & $\gamma$ & $a$ / scalar & -- & $s_a = \gamma$ & \\[\rowsep]
OP58 & \textcode{v3[5]=C2} & -- & $\gamma$ & $a$ / vector & $i$ & $\vec{v}_a^{\,(i)} = \gamma$ & \\[\rowsep]
OP59 & \textcode{m2[5,1]=C1} & -- & $\gamma$ & $a$ / matrix & $i$, $j$ & $M_a^{(i,j)} = \gamma$ & \\[\rowsep]
OP60 & \textcode{s4=uniform(C2,C3)} & -- & $\alpha$, $\beta$ & $a$ / scalar & -- & $s_a = \mathcal{U}(\alpha,\beta)$ & \\[\rowsep]
OP61 & \textcode{m2=m4} & $a$ / matrix & -- & $b$ / matrix & -- & $M_b = M_a$ \\[\rowsep] 
OP62 & \textcode{v2=v4} & $a$ / vector & -- & $b$ / vector & -- & $\vec{v}_b = \vec{v}_a$ \\[\rowsep] 
OP63 & \textcode{i2=i4} & $a$ / index & -- & $b$ / index & -- & $i_b = i_a$ \\[\rowsep] 
OP64 & \textcode{v2=power(v5,v3)} & $a$,$b$ / vectors & -- & $c$ / vector & -- & $\vec{v}_c^{\,(i)} = \operatorname{power}(\vec{v}_a^{\,(i)}, \vec{v}_b^{\,(i)}) \sepforall i$  & \\[\rowsep] 
OP65 & \textcode{v3=m2[:,1]} & $a$,$b$ / matrix,index & -- & $c$ / vector & -- & $\vec{v}_c = M_a^{(\cdot,j_b)}$ & \\[\rowsep] 
OP66 & \textcode{v3=m2[1,:]} & $a$,$b$ / matrix,index & -- & $c$ / vector & -- & $\vec{v}_c = M_a^{(i_b,\cdot)}$ & \\[\rowsep] 
OP67 & \textcode{s3=m2[1,5]} & $a$,$b$,$c$ / m,i,i & -- & $d$ / scalar & -- & $s_d = M_a^{(i_b,j_c)}$ & \\[\rowsep] 
OP68 & \textcode{s3=v2[5]} & $a$,$b$ / vector,index & -- & $c$ / scalar & -- & $s_c = \vec{v}_a^{\,(i_b)}$ & \\[\rowsep] 
OP69 & \textcode{v3=0} & -- & -- & $a$ / vector & -- & $\vec{v}_a = 0$ & \\[\rowsep] 
OP70 & \textcode{s5=0} & -- & -- & $a$ / scalar & -- & $s_a = 0$ & \\[\rowsep] 
OP71 & \textcode{i2=0} & -- & -- & $a$ / index & -- & $i_a = 0$ & \\[\rowsep] 
OP72 & \textcode{v2=sqrt(v5)} & $a$ / vector & -- & $b$ / vector & -- & $\vec{v}_b^{\,(i)} = \operatorname{sqrt}(\vec{v}_a^{\,(i)}) \sepforall i$  & \\[\rowsep] 
OP73 & \textcode{v2=power(v5,2)} & $a$ / vector & -- & $b$ / vector & -- & $\vec{v}_b^{\,(i)} = \operatorname{power}(\vec{v}_a^{\,(i)}, 2) \sepforall i$  & \\[\rowsep] 
OP74 & \textcode{s1=sum(v5)} & $a$ / vector & -- & $b$ / scalar & -- & $s_b = \operatorname{sum}(\vec{v}_a^{\,(i)}) \sepforall i$  & \\[\rowsep] 
OP75 & \textcode{s5=sqrt(s1)} & $a$ / scalar & -- & $b$ / scalar & -- & $s_b = \sqrt{s_a}$ & \\[\rowsep] 
OP76 & \textcode{s3=s0*s2+s5} & $a$,$b$,$c$ / scalars & -- & $d$ / scalar & -- & $s_d = s_a*s_b+sc$ & \\[\rowsep] 
OP77 & \textcode{s2=s4*C1} & $a$ / scalar & $\gamma$ & $b$ / scalar & -- & $s_b = s_b * \gamma$ & \\[\rowsep] 
OP78 & \textcode{m2[1,:]=v3} & $a$ / vector & -- & $b$ / matrix & $i$ & $M_b^{(i,\cdot)} = \vec{v}_a$ & \\[\rowsep] 
OP79 & \textcode{m2[:,1]=v3} & $a$ / vector & -- & $b$ / matrix & $i$ & $M_b^{(\cdot,j)} = \vec{v}_a$ & \\[\rowsep] 
OP80 & \textcode{i3 = size(m1, axis=0) - 1} & $a$ / matrix & -- & $b$ / index & -- & $i_b = \operatorname{size}(M_a^{(i,\cdot)}) - 1$ & \\[\rowsep] 
OP81 & \textcode{i3 = size(m1, axis=1) - 1} & $a$ / matrix & -- & $b$ / index & -- & $i_b = \operatorname{size}(M_a^{(\cdot,j)}) - 1$ & \\[\rowsep] 
OP82 & \textcode{i3 = len(v1) - 1} & $a$ / vector & -- & $b$ / index & -- & $i_b = \operatorname{len}(\vec{v}_a) - 1$ & \\[\rowsep] 
OP83 & \textcode{s1 = v0[3] * v1[3] + s0} & $a$,$b$,$c$,$d$ / v,v,s,i & -- & $e$ / scalar & -- & $s_e =  \vec{v}_a^{\,(i_d)} *  \vec{v}_b^{\,(i_d)} + s_c$& \\[\rowsep] 
OP84 & \textcode{s3=dot(v0[:5],v1[:5])} & $a$,$b$,$c$ / v,s,i & -- & $d$ / scalar & -- & $s_d=\vec{v}_a^{\,T(:i_c+1)} \, \vec{v}_b^{\,(:i_c+1)}$ & \\[\rowsep] 
\cmidrule(r{8pt}){1-7}
\end{tabular}
\end{scriptsize}
\end{center}
\end{table*}

\end{document}